\documentclass[preprint,12pt]{elsarticle}

\usepackage{hyperref}
\usepackage{float}
\usepackage{multirow}
\usepackage{booktabs}
\usepackage{array}
\usepackage{tabularx}
\newcolumntype{L}[1]{>{\raggedright\arraybackslash}p{#1}}
\newcolumntype{C}{>{\centering\arraybackslash}X}
\usepackage{float}
\usepackage{amssymb}
\usepackage[a4paper, top=2.5cm, bottom=2.5cm, left=1.9cm, right=1.7cm]{geometry}
\usepackage{amsmath}
\usepackage{cleveref}
\usepackage{soul} 
\usepackage[normalem]{ulem} 
\usepackage{algorithm}
\usepackage{algpseudocode}
\newcommand\redsout{\bgroup\markoverwith
{\textcolor{blue}{\rule[0.5ex]{2pt}{0.4pt}}}\ULon}

\usepackage{xcolor}
\usepackage{graphicx}
\newcommand{\RevisionBlue}[1]{{\color{black}#1}}

\usepackage{threeparttable} 

\definecolor{myblue}{RGB}{0, 100, 200} 
\newcommand{\AM}[1]{\textcolor{black}{#1}}

\journal{Elsevier}

\begin{document}

\begin{frontmatter}

\title{Deep learning-based prediction of time-resolved adhesive forces in viscoelastic Hertzian contacts}

\author[inst2,inst1]{Ali Maghami\corref{cor1}}
\ead{ali.maghami@tu-berlin.de}
\cortext[cor1]{Corresponding author}

\affiliation[inst2]{organization={Chair of Cyber-Physical Systems in Mechanical Engineering, Technische Universität Berlin},
            addressline={Straße des 17. Juni}, 
            city={Berlin},
            postcode={10623},
            country={Germany}}

\affiliation[inst1]{organization={TriboDynamics Lab, Department of Mechanics, Mathematics and Management, Polytechnic University of Bari},
            addressline={Via Orabona 4}, 
            city={Bari},
            postcode={70125}, 
            country={Italy}}

\author[inst2]{Merten Stender}
\author[inst1]{Michele Ciavarella}
\author[inst1]{Antonio Papangelo}

\begin{abstract}

Fast prediction of the response of adhesive soft viscoelastic contacts represents a current challenge in soft robotics and for gripping and manipulation tasks. In viscoelastic adhesive contacts, the loading-dwell-unloading protocol has a tremendous impact on the maximum adherence force, work of separation, and time of the pull-off event of the soft interface. Determining the complete time-resolved force trajectory requires full numerical simulations, whose computational cost is strongly parameter-dependent, making them impractical for real-time identification or design-optimization loops. In this work, we overcome this limitation by training a scalar-conditioned, stateful, sequence-to-sequence deep learning model to predict the full force evolution from a prescribed displacement history for both short- and long-range adhesion regimes. A boundary element solver, combining a viscoelastic substrate with a Lennard-Jones traction-separation law, generated \AM{12,446} dimensionless force trajectories used to train, evaluate, and test the proposed approach. The data set spans four orders of magnitude in loading and unloading rates and includes varied dwell times, with the Tabor parameter ranging from $0.2$ to $3.2$. To enable learning across these heterogeneous time scales, we introduce a fixed-measurement-step (FMS) representation that converts variable-length trajectories into fixed-length sequences while preserving their physical-time information. Eighteen different architectures were trained, including long short-term memory (LSTM) networks, temporal convolutional neural (TCN) networks, and time-distributed dense layers with three different Tabor-conditioning mechanisms. The models were compared using global waveform and error metrics. We found that the best-performing model has an LSTM architecture with concatenated conditioning, which achieves a held-out mean-squared error of $5.0\times10^{-4}$, a median pull-off-force error of $\approx2.2\%$, and a median hysteresis error of $\approx1.1\%$. For the held-out protocols, the model predicts a complete force trajectory with a median inference time of $0.16$ s. \RevisionBlue{An illustrative analysis of total computational cost further identifies combinations of query volume and acceptable mean test error for which direct BEM or the trained surrogate is more cost-effective.} The model is tested across unseen parameter combinations and against analytical limiting cases, providing a rapid surrogate for repeated numerical evaluations with potential use in control-oriented applications. All codes and data are shared openly on GitHub and Zenodo for further development and reuse.

\end{abstract}

\begin{graphicalabstract}
\includegraphics[width=1\textwidth]{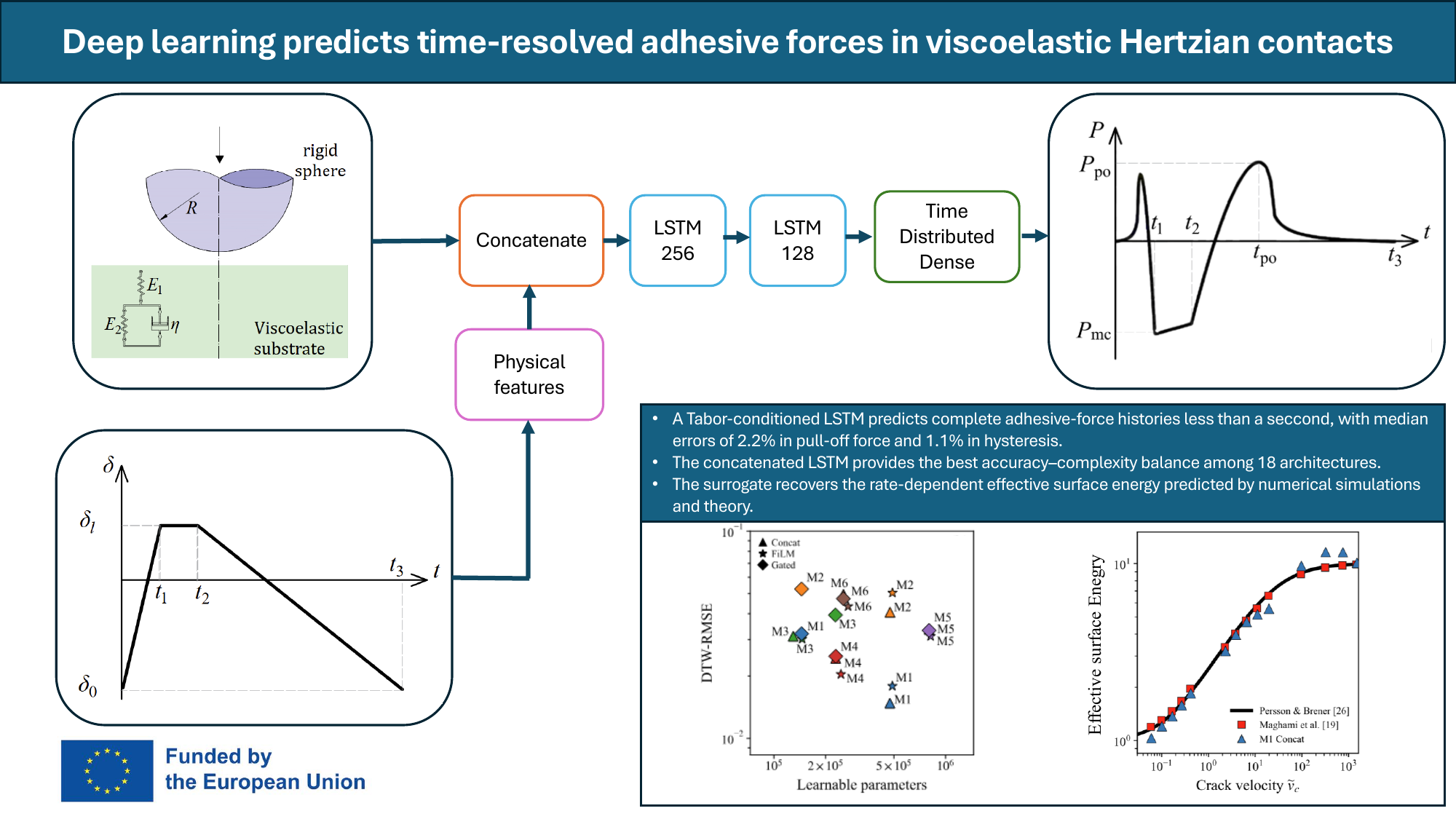}
\end{graphicalabstract}

\begin{highlights}
    \item {A seq2seq LSTM predicts time-resolved adhesive force in viscoelastic contact.}
    \item {Eighteen architectures compared; LSTM achieves 0.05 \% mean-squared error.}
    \item {Fast parameter-independent inference time of about 0.25 s for a full force branch.}
    \item {Fixed measurement-step representation resolves heterogeneous sequence lengths.}
    \item {Fast pull-off and hysteresis prediction with controlled error.}
    
\end{highlights}

\begin{keyword}
Viscoelastic adhesion \sep Hertzian contact \sep Deep learning \sep Time-resolved \sep Adhesive forces \sep Sequence-to-sequence modeling \sep Surrogate model

\end{keyword}

\end{frontmatter}



\section{Introduction}

\begin{figure}[htbp]
    \centering
\includegraphics[width=0.85\textwidth]{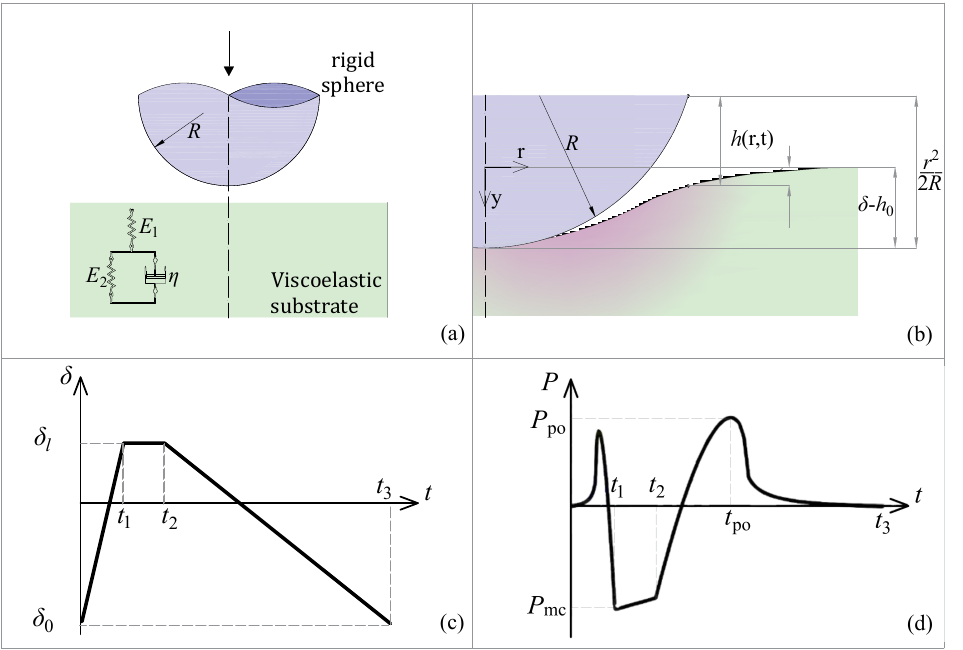}

    \caption{Schematic of indentation of a viscoelastic substrate by a rigid spherical indenter: (a) the sphere, substrate and loading configuration, (b) geometric components of the gap function $h(r,t)$ at an arbitrary contact point, including the indentation $\delta$, the equilibrium spacing $h_0$, the parabolic indenter profile $r^2/(2R)$, and the viscoelastic surface deformations $u_z(r,t)$ which depend on the time $t$; (c) displacement-based loading-dwell-unloading protocol, with the initial and final indentation $\delta_0$ chosen sufficiently far from the contact interface that the adhesive force is negligible at both sequence endpoints; (d) representative adhesive force response $P(t)$ during the loading-dwell-unloading protocol, indicating the pull-off force $P_{\mathrm{po}}$, pull-off time $t_{\mathrm{po}}$, and maximum compressive load $P_{\mathrm{mc}}$.}
    \label{fig:load_and_unloading}
\end{figure}

The contact between a soft adhesive body and a rigid indenter under a prescribed loading-dwell-unloading cycle produces a time-resolved force trajectory that encodes substantially more information than the peak adherence force alone. The enclosed hysteresis loop as shown in Figure \ref{fig:load_and_unloading}, determines the energy dissipated at the interface, the shape of the unloading branch governs the stability of the contact under varying rates, and the precise timing of the snap-off event is critical for controlled release in gripping, adhesion-switching applications and docking capabilities \cite{ji2019apple, liu2022switchable, liang2024autopeel, ziemer2025impact}. Rapid prediction of this complete force history is essential for closed-loop control of soft robotic end-effectors \cite{giordano2024mechanochromic, tao2023climbing}, for iterative material-design evaluations \cite{humfeld2021machine}, and for dynamic biomechanical simulations \cite{guo2024interaction}. Reproducing the force trajectory at a computational cost compatible with repeated real-time evaluation is the central challenge addressed in this work.

The theoretical framework for adhesive Hertzian contact rests on two well-established elastic limiting solutions. The Johnson-Kendall-Roberts (JKR) theory \cite{johnson1971surface}, valid for soft compliant contacts governed by short-range surface forces, and the Derjaguin-Muller-Toporov (DMT) model, applicable to stiff contacts with long-range adhesion \cite{derjaguin1975effect}, predict pull-off forces of $(3/2)\pi R\Delta\gamma_0$ and $2\pi R\Delta\gamma_0$, respectively, where $R$ is the sphere radius and $\Delta\gamma_0$ is the thermodynamic surface energy \cite{PAPANGELO2026}. The transition between these two limiting regimes is governed by the Tabor parameter \cite{tabor1977surface} $\mu=\left( \frac{R \Delta\gamma_0^2}{{E^*}^2 h_0^3} \right)^{1/3}$, a dimensionless ratio between the height of the neck at the periphery of the contact patch and the range of surface forces $h_0$, being $E^*=E/(1-\nu^2)$ with $E$ the Young's modulus and $\nu$ the Poisson's ratio, such that large $\mu$ corresponds to JKR-like and small $\mu$ to DMT-like behavior \cite{PAPANGELO2026, maugis1992adhesion, violano2021jkr}. Intermediate values of $\mu$ require the introduction of a cohesive zone model, which was first introduced by Maugis \cite{maugis1992adhesion} who showed that in the limit of large and small Tabor parameters, the JKR and DMT solutions can be retrieved, respectively. Crucially, these classical models consider the contacting material as elastic; hence, they are rate-independent, and the pull-off force is set by the geometry and surface energy regardless of how the pull-off state was reached.

The adhesive response of soft polymers and elastomers departs fundamentally from this rate-independent picture. Bulk viscoelastic dissipation amplifies the effective (often termed apparent) surface energy $\Delta\gamma_{\mathrm{eff}}$ above the thermodynamic baseline \cite{PAPANGELO2026, tricarico2025enhancement,tricarico2026influence, ciavarella2025dynamic, maghami2024viscoelastic, maghami2024bulk, violano2021jkr, afferrante2022effective, carbone2022theory, mandriota2024adhesive, shui2020rapid}, and the instantaneous contact force depends on the entire preceding loading path rather than on the current contact state alone \cite{tricarico2026influence,afferrante2022effective}; the amplification factor of the adherence force approaches the ratio $E_\infty/E_0$ which can reach several orders of magnitude in silicone-based polymers, being $E_\infty$ the instantaneous and $E_0$ the relaxed Young modulus \cite{schapery1975theory1,greenwood1981mechanics,persson2005crack,maghami2024bulk}. Analytical theories have addressed this problem in specific limiting regimes. Cohesive-zone models \cite{schapery1975theory1, greenwood2004theory} account for the rate-dependent process zone size at the contact edge (the crack "mouth"), while energy-based theories \cite{persson2005crack,carbone2022theory} establishes an energy balance for the propagating contact front. Closed form analytical solutions exist for predicting $\Delta\gamma_{\mathrm{eff}}$ based on the proposed theories \cite{schapery1975theory1, greenwood2004theory,persson2005crack}, having the latter being extended also for viscoelastic materials with wide-band spectrum \cite{maghami2024bulk}, nevertheless the latter approaches are restricted to quasi-static detachment from a fully relaxed initial state, apply in the JKR-like adhesion limit, and yield at most a scalar effective surface energy rather than a complete force trajectory. As Johnson observed, viscoelastic adhesion remains ``\textit{a difficult problem in contact mechanics}'' \cite{popov2021note}, and no closed-form model provides the full force evolution under an arbitrary loading-dwell-unloading history; hence, the solution is restricted to numerical approaches \cite{maghami2024bulk, papangelo2023detachment,carbone2022theory,mandriota2024adhesive}.

\RevisionBlue{Numerical simulation, together with analytical methods, is central to computational contact mechanics. Recent computational-mechanics studies extend contact modeling to nonlinear mechanical response and nonclassical material behavior, combining analytical, finite-element, and machine-learning approaches for functionally graded or orthotropic contacts and examining adhesion in piezoelectric, thermoelectric, or quasicrystalline materials \cite{revAbushattal2026,revZhang2026,revZhang2025,revDuan2025,revMa2025,revYaylaci2025,revOner2025}. For viscoelastic adhesive contact, a boundary element method (BEM) coupled with a time-marching Newton--Raphson scheme} \cite{ahmad2024family} can compute the complete force trajectory across loading protocols, Tabor parameters, and initial conditions \cite{papangelo2020numerical, papangelo2023detachment}. The computational cost, however, is substantial and strongly input-dependent. Hence, simulation time scales as $\mathcal{O}(N_s^2\,N_t\,N_\mathrm{iter})$, where $N_s$ is the number of spatial nodes, $N_t$ the number of time steps, and $N_\mathrm{iter}$ the average nonlinear iteration count per step, and it grows sharply with increasing Tabor parameter (which requires finer spatial discretization), increasing indentation depth, and decreasing unloading rate \cite{maghami2024bulk, souza2010multiscale}. Across the parameter space considered in this work, individual simulations range over more than three orders of magnitude on the same computational platform. This variability arises from the coupled effects of adaptive time-step refinement, spatial discretization requirements that scale with contact-zone extent, and nonlinear-iteration counts that increase with adhesive-zone complexity.

Machine learning and deep learning (DL) offer fast surrogates with fixed inference cost once trained, irrespective of the complexity encoded in the input parameters \cite{guo2021artificial, mackay2023informed}. These methods have demonstrated broad potential in materials science \cite{lu2026llm, javadi2022deep, kellner2019establishing, eshkofti2024modified, yan2021machine}, fracture mechanics \cite{wang2021machine, athanasiou2023integrated, yi2024mechanics, perera2023generalized, li2022machine}, \RevisionBlue{contact mechanics \cite{goodbrake2024neural, motiwale2024neural, kalliorinne2021artificial, sahin2024solving, sahin2024physics, revYaylaci2025, revOner2025},} and tribology \cite{stender2021deep, geier2023machine, sattari2020prediction}. Within adhesion, existing studies have concentrated on the geometric optimization of fibrillar and micropatterned contacts \cite{kim2020designing, son2021machine, luo2022machine, kim2023designing, dayan2024machine, shojaeifard2025machine} or on the prediction of scalar detachment quantities for flat-to-flat contact configurations \cite{nguyen2026machine}. A first step toward Hertzian viscoelastic adhesion was taken by Maghami et al. \cite{maghami2025pull}, who developed a physics-augmented tree-based machine learning framework for predicting the pull-off force and work-to-pull-off across wide ranges of the Tabor parameter, material properties, and unloading rate. 

\RevisionBlue{These abstract detachment metrics, however, are aggregate outcomes of the force--displacement response and do not explicitly capture the path-dependent force trajectory produced by viscoelastic memory during loading, dwell, and unloading. Viewed against the broader analytical, numerical, and data-driven contact-mechanics literature \cite{revAbushattal2026,revZhang2026,revZhang2025,revDuan2025,revMa2025,revYaylaci2025,revOner2025}, the contribution of the present work is not simply the application of deep learning; rather, it is the prediction of the complete time-resolved adhesive force trajectory under loading--dwell--unloading histories. This trajectory-level objective goes beyond predicting only scalar quantities such as maximum pull-off force, work of separation, or effective surface energy \cite{maghami2025pull,nguyen2026machine,schapery1975theory1,greenwood2004theory,persson2005crack,carbone2022theory}.}

A key structural distinction separates this prediction task from abstract surrogate modeling. Viscoelastic adhesive contact is governed by a Boltzmann convolution integral (see Eq.~\eqref{eq:uz}): hence, it cannot be recovered from the instantaneous contact state alone. This motivates stateful sequence models such as LSTMs \cite{hochreiter1997long}, which update an internal state along the loading history to predict the full force trajectory.
Precedents for this approach include LSTM architectures that learn constitutive viscoelastic behavior from stress-strain histories \cite{chen2021recurrent}, physically consistent formulations through structured network design \cite{koeppe2022explainable}, and effective surrogates for history-dependent contact and manipulation tasks \cite{george2022closing, wang2024drift, li2023modeling, karami2023real, hinrichsen2024using}.

Unlike the preceding work of Maghami et al.\ \cite{maghami2025pull}, which employed tree-based tabular models to predict abstract detachment quantities for contacts unloaded from a fully relaxed initial state, the present study targets the prediction of the full time-resolved adhesive force trajectory, including contact states that are not relaxed at the onset of unloading. The surrogate is therefore required to reproduce the loading branch, the dwell-induced relaxation transient, the complete unloading path, and the timing of the snap-off event from the prescribed displacement history and the Tabor parameter alone. This task cannot be reduced to a static input-output mapping and directly motivates the use of stateful deep sequence models. The constitutive description is restricted to a single-relaxation viscoelastic model so that this first sequence-prediction benchmark for viscoelastic adhesive contact can be posed, trained, and evaluated in a controlled and reproducible setting before extension to broader material spectra \cite{maghami2024bulk,williams1964structural,mainardi2011creep}.

The objectives of this work are: (i) to introduce a fixed measurement-step representation to accommodate the heterogeneous time-sequence lengths arising from a broad numerical dataset spanning a variation of several orders of magnitude in terms of loading and unloading rates, indentation, dwell times for varying Tabor parameter; (ii) to perform an extensive architecture search over scalar-conditioned sequence-to-sequence models, spanning six family networks and three Tabor-conditioning mechanisms, to identify design choices that can inform future trajectory-level surrogates for history-dependent adhesive contact; (iii) to assess the role of physics-guided input features by contrasting a minimally processed input configuration with an augmented representation incorporating causal velocity and protocol edge-detection channels; (iv) to validate the reference surrogate against numerical simulations using physically interpretable detachment quantities, including pull-off force, pull-off time, and hysteresis also against analytical limiting case through effective surface energy; and (v) to provide a rich, first-of-its-kind dataset of time-resolved viscoelastic adhesive contact trajectories for future benchmarking and developments. \RevisionBlue{We also combine the observed test errors for different training fractions with an amortized-cost analysis to identify combinations of accuracy requirement and query volume that favor direct BEM or the DL+BEM framework under the stated cost assumptions.} The remainder of this paper is structured as follows: Section~\ref{sec:methods} describes the data-generation procedure, the sequence-model architectures, and the evaluation metrics; Section~\ref{sec:results} presents the prediction results; Section~\ref{sec:discussion} discusses the key findings; and Section~\ref{sec:conc} summarizes the main conclusions.

\section{Methods}\label{sec:methods}

\subsection{Physical model}\label{sec:physmodel}

Let us consider a rigid sphere of radius $R$ pressed against and subsequently retracted from an adhesive viscoelastic half-space (Figure~\ref{fig:load_and_unloading}), with the initial and final indentation chosen sufficiently remote from the interface that the adhesive force is negligible at both sequence endpoints. We restrict our attention to ramp loading and unloading performed respectively at the constant velocities $v_L$ and $v_U$ (Figure~\ref{fig:load_and_unloading}c). The Standard Linear Solid (SLS) is considered as a classical viscoelastic material model, constituted by a spring in parallel with a dashpot and the pair in series with a spring, as it reproduces the essential features of a viscoelastic material. The interactions between the rigid indenter and the viscoelastic halfspace are described by the Lennard-Jones traction-separation law 
\cite{ciavarella2019role, johnson1997adhesion}:

\begin{equation}
\sigma(h) = \frac{8\Delta\gamma_{0}}{3h_{0}} \left[ \left(\frac{h_{0}}{h}\right)^{3} - \left(\frac{h_{0}}{h}\right)^{9} \right],
\label{eq:LJ}
\end{equation}
where $\sigma$ is the interfacial stress (positive when tensile), $h$ is the local gap between surfaces, and $h_{0}$ is the equilibrium separation distance at which $\sigma(h_0) = 0$. The thermodynamic surface energy $\Delta\gamma_{0}$ is related to the maximum tensile stress $\sigma_0$, occurring at $h = 3^{1/6}h_0$, by $\Delta\gamma_{0} = \frac{9\sqrt{3}}{16} \sigma_{0} h_{0}$ \cite{greenwood1997adhesion}.

The gap function $h(r,t)$ between the rigid sphere and the substrate at radial coordinate $r$ and time $t$ is:
\begin{equation}
h(r,t) = -\delta(t) + h_{0} + \frac{r^{2}}{2R} + u_{z}(r,t),
\label{eq:h}
\end{equation}
where $\delta(t)$ is the indentation depth (positive as the sphere moves towards the substrate), and $u_{z}(r,t)$ is the viscoelastic surface displacement (positive in the half-space), which depends on the full loading history \cite{christensen2012theory}. Applying the elastic-viscoelastic correspondence principle via Boltzmann superposition \cite{greenwood1997adhesion, christensen2012theory,feng2000contact}:
\begin{equation}
u_{z}(r,t) = \int_{0}^{\infty} G(r,s)\, s \int_{-\infty}^{t} C(t - \tau) \frac{\partial\sigma(s, \tau)}{\partial\tau}\, d\tau\, ds,
\label{eq:uz}
\end{equation}
where $G(r,s)$ is the elastic Kernel function of the half-space (given explicitly in \ref{app:BEM_disc}) and $C(t)$ is the creep compliance function. For the SLS model adopted here the creep compliance function is
\begin{equation}
C(t) = \frac{1}{E_0} \left[ 1 + (k - 1)\exp\!\left(-\frac{t}{\tau_r}\right) \right],
\label{eq:Ctgen}
\end{equation}
where $\tau_r$ is the retardation time, $k = E_0/E_\infty$ is the modulus ratio, and $E_0$, $E_\infty$ are the rubbery and glassy moduli, respectively.

Equations~(\ref{eq:LJ}) to (\ref{eq:Ctgen}) are combined into a nonlinear convolution integral equation for the unknown gap field $h(r,t)$, whose numerical solution is carried out in dimensionless form. The nondimensional parameters are defined as
\begin{equation}
\widehat{\delta} = \frac{\delta}{h_0}, \quad
\widehat{t} = \frac{t}{\tau_r}, \quad
{\widehat{v}=\frac{v\tau_r}{h_0}}, \quad
\widehat{P}=\frac{P}{\pi\Delta\gamma_0 R},
\label{eq:dlesspar}
\end{equation}
where $\widehat{P}$ is the dimensionless normal force. The Tabor parameter $\mu=\left( \frac{R \Delta\gamma_0^2}{{E^*_0}^2 h_0^3} \right)^{1/3}$ governs the transition between the long-range ($\mu \lesssim 0.2$) and short-range ($\mu  \gtrsim 3$) adhesion regimes, and it appears as the key non-geometric input to the surrogate models.

The dimensionless governing equations are solved numerically by BEM combined with a Newton-Raphson iteration on $N_s$ equally spaced nodes \cite{papangelo2020numerical,papangelo2023detachment, maghami2025pull}. Temporal discretization proceeds by a time-marching algorithm with step $\Delta\widehat{t}$, and spatial discretization employs the method of overlapping triangles \cite{johnson1987contact}, which assumes a piecewise-linear pressure distribution over each element. Details of the discretized equations, the influence matrix, and the dimensionless forms of Eqs.~(\ref{eq:LJ}), (\ref{eq:h}), and (\ref{eq:uz}) are given in \ref{app:BEM_disc}. The results obtained with this numerical code have been validated for both flat and Hertzian viscoelastic adhesive contacts in Ref.s \cite{papangelo2023detachment, maghami2025pull}.   
\subsection{Dataset generation}\label{sec:dataset}

To systematically explore the adhesive response of the indenter-halfspace contact, five parameters were varied: the loading rate $\widehat{v}_L=\frac{{\widehat{\delta}_l}-\widehat{\delta}_0}{\widehat{t}_1}$ and unloading rate $\widehat{v}_U=\frac{\widehat{\delta}_l-\widehat{\delta}_0}{\widehat{t}_3 -\widehat{t}_2}$(Figure~\ref{fig:load_and_unloading}(c)), each value independently drawn over $[10^{-1},\,10^{3}]$, the dwell time $\widehat{t}_D=\widehat{t}_2-\widehat{t}_1$ over $[10^{-3},\,3]$, indentation depth $\widehat{\delta_l}$ over [0, 100] and the Tabor parameter $\mu$ over $[0.2,\,3.2]$. We note that the initial and final indentation ${\delta}_0$ for all the samples are chosen sufficiently far from contact, equal to $-{\pi^{2/3}\mu}$. Lower loading rates produce rubbery, compliant responses; higher rates produce glassy, stiff ones. Short dwell times correspond to near-instantaneous unloading after the loading is completed, while longer dwell times allow stress relaxation. The Tabor parameter spans the regimes from long-range ($\mu = 0.2$) to short-range ($\mu = 3$) adhesive interaction \cite{greenwood1997adhesion}. The modulus ratio was fixed at $k = 0.1$. A total of \AM{12,446} trajectories were generated from this parameter space. 
\begin{figure}[htbp]
    \centering
\includegraphics[width=0.32\linewidth]{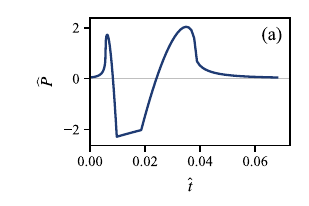}
\includegraphics[width=0.32\linewidth]{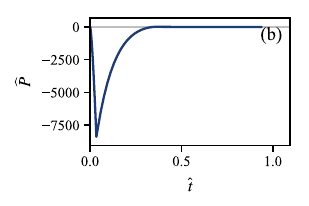}
\includegraphics[width=0.32\linewidth]{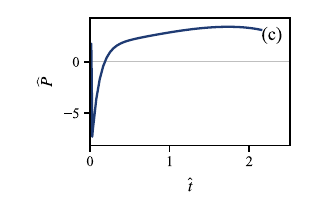}\\
\includegraphics[width=0.32\linewidth]{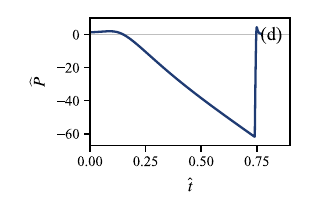}
\includegraphics[width=0.32\linewidth]{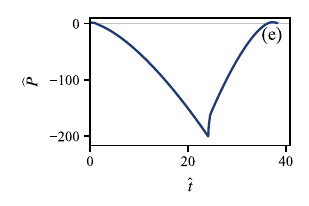}
\includegraphics[width=0.32\linewidth]{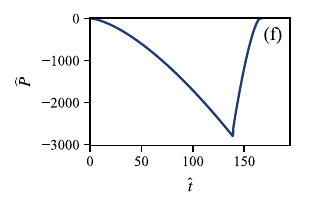}\\
\includegraphics[width=0.32\linewidth]{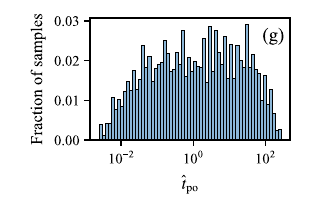}
\includegraphics[width=0.32\linewidth]{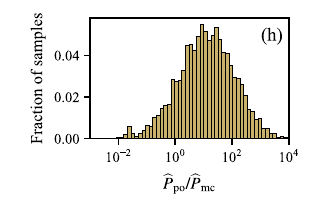}
\includegraphics[width=0.32\linewidth]{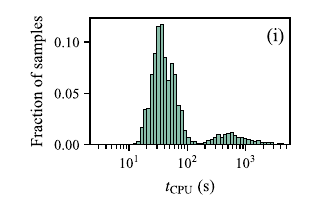}
\caption{Representative force-time responses and empirical statistics for the simulated dataset. Panels (a)--(f) show six selected trajectories on linear time axes, spanning short, intermediate, and long pull-off times together with weak and strongly negative pull-off forces. The bottom row contains three histograms normalized by the total number of simulation considered: (g) pull-off time $\widehat{t}_\textrm{po}$ on a logarithmic axis, (h) force-range ratio $\widehat{P}_\textrm{po}/\widehat{P}_\textrm{mc}$ on a logarithmic axis, and (i) BEM simulation time $t_{CPU}(\textrm{s})$ distribution on a logarithmic axis, showing computational cost variability spanning more than three orders of magnitude. Loading-protocol parameters for panels (a)--(f) are listed in Table~\ref{tab:fig2_protocol_parameters} in ~\ref{sec:appendix_fig2}.}
    \label{fig:force_vs_time_unsteady}
\end{figure}

Figure~\ref{fig:force_vs_time_unsteady} summarizes the dataset variability using six representative force-time trajectories (panels (a-f)) together with the frequency plots of the dimensionless time at which pull-off happens $\widehat{t}_{po}$ (panel (g)), the pull-off-to-maximum-preload ratio $\widehat{P}_{\mathrm{po}}/|\widehat{P}_{\mathrm{mc}}|$ (panel (h)), where $\widehat{P}_{\mathrm{po}}$ is the pull-off force and $|\widehat{P}_{\mathrm{mc}}|$ is the magnitude of the maximum compressive load reached in the same trajectory (see Figure \ref{fig:force_vs_time_unsteady}d)
, and BEM simulation time $t_\textrm{CPU}$ (Figure \ref{fig:force_vs_time_unsteady} (i)). In the investigated parameter space the dimensionless time at which the pull-off event may occur can vary over about 5 orders of magnitude (panel~(g)), from very rapid snap-offs to cases extending to nearly $\widehat{t}_\mathrm{po}\approx3\times10^{2}$, while the associated force response spans from a small positive regime prior to pull-off to markedly negative force values in compression resulting in a distribution of $\widehat{P}_{\mathrm{po}}/|\widehat{P}_{\mathrm{mc}}|$ that spans 5 orders of magnitude. For the learning surrogate, this means that it has to learn the trajectory while the pull-off event may be comparable to the compressive load scale in some histories but orders of magnitude smaller in others.

As mentioned here and discussed in our previous works \cite{maghami2024bulk, maghami2025pull}, computational cost increases rapidly for large Tabor parameters, deeper indentation, and slower unloading. Individual simulations across the parameter space studied here ranged from a few seconds to hours of computational times, depending on input parameters (panel~(i)), using MATLAB 2023b$^{\text{\copyright}}$ on a desktop computer equipped with Windows 11 Pro, a 12th Gen Intel(R) Core(TM) i9-12900K, 3200 MHz, 16 Cores, and 96 GB RAM. \textcolor{black}{The successful BEM simulations retained in the dataset required approximately 478 single-core CPU-hours (about 20 CPU-days). This count excludes preliminary discretization studies and unsuccessful attempts, so the full effort required to construct the database was larger.} \textcolor{black}{On our workstation, this translates to an average of about 30 hours for single-core usage.} This computational-cost skewness \textcolor{black}{in BEM simulations} underscores the practical motivation for a fixed-cost surrogate, while the BEM solver's runtime scales nonlinearly with problem complexity and discretization requirements, the trained sequence model provides uniform inference time across the entire parameter space. This enables the rapid repeated evaluations required for control, optimization, and inverse-identification applications. Together, these observations imply two modeling challenges: (i) heterogeneous sequence lengths, because each trajectory terminates at a different number of time steps; and (ii) a broad dynamic range in both inputs and outputs, spanning several orders of magnitude. These challenges motivate the representation strategy described in the following subsection.

\subsection{Data mapping}\label{sec:dm}
The two modeling challenges identified above (heterogeneous sequence lengths and broad dynamic range) make direct use of the raw BEM time series impractical for batch learning. Two naive alternatives exist, but are both unsatisfactory. An event-driven approach, which terminates each sequence at pull-off, implicitly informs the model of the pull-off time and can bias predictions. A fixed-time-grid approach, which uses the longest trajectory's discretization for all samples, produces excessively long sequences, even when not needed. This amplifies vanishing-gradient problems and training cost drastically.

We resolve both issues by representing the load-time trajectories with a fixed number of measurement steps, replacing the variable number of time steps. We will refer to this as a Fixed Measurement Step (FMS) representation. The FMS representation assigns a prescribed number of measurement points to each phase of the loading protocol instead of using the native adaptive BEM time discretization directly. In the present dataset, each trajectory is represented by $N = 120$ points, including 50 points for loading, 20 points for dwell, and 50 points for unloading. We note that the points are uniformly distributed within each phase. The proposed representation is conceptually similar to fixed-resolution approaches such as PAA/SAX \cite{lin2003symbolic}, but it imposes a prescribed number of measurement points within each physical phase. Because each BEM simulation is generated on its own adaptive time grid, spline interpolation is used to map the raw trajectory onto this common FMS grid.

Formally, let us consider a BEM-generated numerical trajectory represented as $\{\widehat{t}_{q},\widehat{\delta}_{q},\widehat{P}_{q}\}_{q=1,...,Q}$, being $Q$ the number of time steps for that trajectory before FMS resampling. The FMS mapping $\mathcal{S}$ produces a fixed-length sequence through spline interpolation as follows:

\begin{equation}
\mathcal{S}:\;\bigl\{\widehat{t}_q,\widehat{\delta}_q,\widehat{P}_q\bigr\}_{q=1,...,Q}
\;\longmapsto\;
\bigl\{\widehat{t}_j,\widehat{\delta}_j,\widehat{P}_j\bigr\}_{j=1,...,N}\;
\end{equation}
where, $j=1,\dots,N$ labels the fixed measurement steps after resampling. The FMS index ranges $j=1,\ldots,N_L$, $j=N_L+1,\ldots,N_L+N_D$, and $j=N_L+N_D+1,\ldots,N$ preserve the loading, dwell, and unloading portions of the original BEM trajectory, while spline interpolation maps the adaptive BEM samples within each phase onto the prescribed number of measurement steps, being $(N_L,N_D,N_U)=(50,20,50)$.

\begin{figure}[h]
    \centering
    \includegraphics[width=0.92\linewidth]{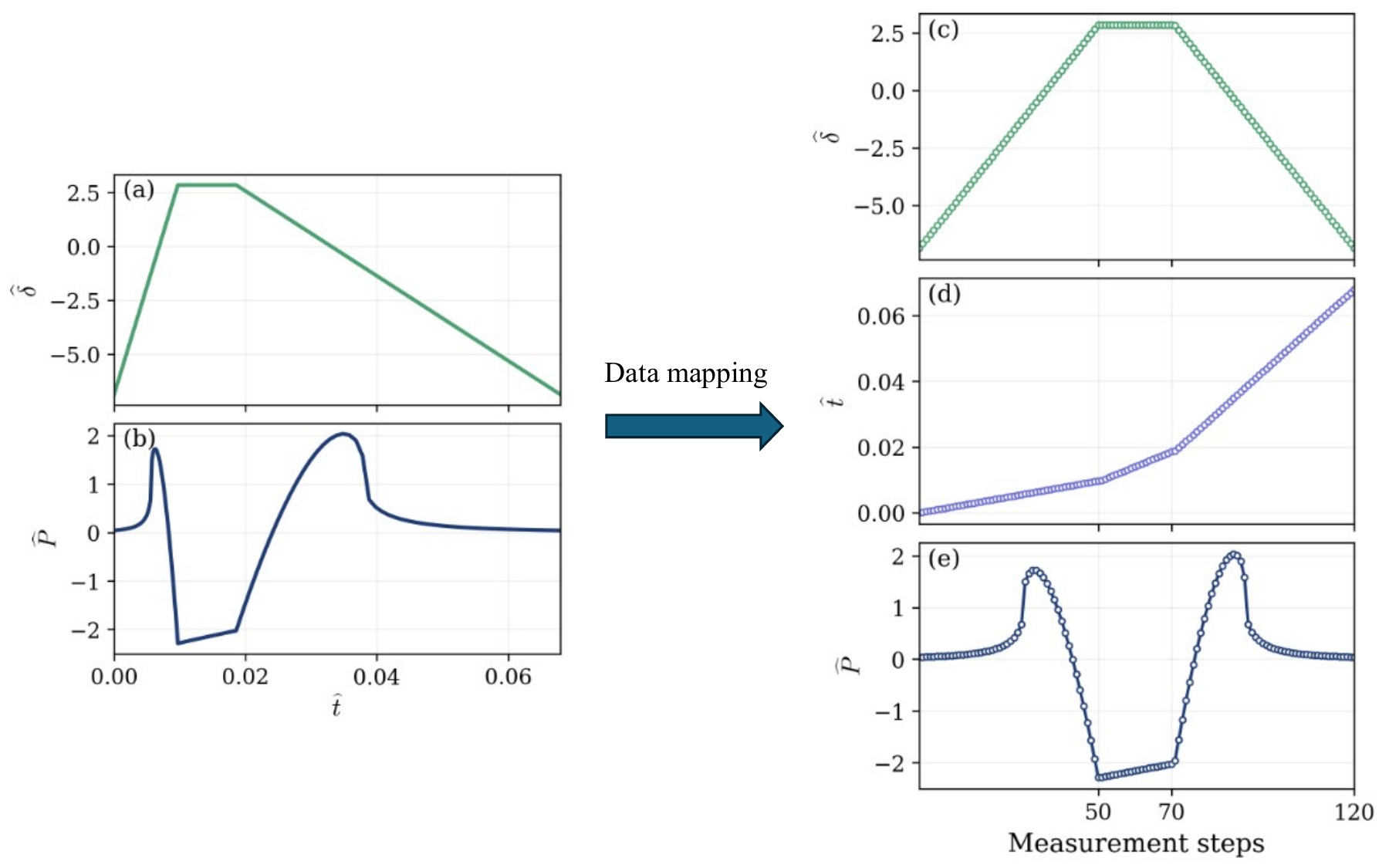}
    \caption{Fixed-measurement-step (FMS) mapping from the native BEM time series to a fixed-length sequence with $N=120$ measurement steps. Panel~(a) shows the normalized indentation history $\widehat{\delta}$ as a function of normalized time $\widehat{t}$ on the original time axis. Panel~(b) shows the corresponding normalized force response $\widehat{P}$ as a function of $\widehat{t}$. Panel~(c) shows the same indentation history after resampling onto the FMS measurement index. Panel~(d) shows the mapped normalized time channel $\widehat{t}$ associated with each measurement step. Panel~(e) shows the mapped normalized force target $\widehat{P}$ on the same fixed measurement grid.}
    \label{fig:input_output}
\end{figure}

Figure~\ref{fig:input_output} visualizes the FMS encoding after the preprocessing steps described above. The original indentation and force trajectories are first expressed on the native time axis, then remapped to the common measurement-step grid used by the sequence model. The mapped channels retain the loading, dwell, and unloading structure while giving every trajectory the same sequence length.

\subsection{Surrogate-model inputs and targets}
The surrogate receives two qualitatively different inputs. The first is dynamic and sequential, the FMS-resampled loading history, stored in a matrix $\mathbf{X}$ whose rows follow the measurement index $j=1,\ldots, N$. The second is static, the Tabor parameter $\mu$, which is fixed for the whole trajectory and specifies the adhesion regime. The target is also sequential, namely the normalized force history $\mathbf{y}$. With $N=120$, the learning problem is written as
\begin{equation}
{
\begin{aligned}
\mathbf{X}
&=\left[
\mathbf{x}_1,\mathbf{x}_2,...,\mathbf{x}_d\right]
\in\mathbb{R}^{N\times d},
\qquad
\mu\in\mathbb{R},
\qquad
\mathbf{y}
=
(\widehat{P}_1,\ldots,\widehat{P}_N)^{\top}
\in\mathbb{R}^{N},\\
\widehat{\mathbf{y}}
&=
f_{\theta}(\mathbf{X},\mu),
\qquad
f_{\theta}:\mathbb{R}^{N\times d}\times\mathbb{R}\to\mathbb{R}^{N}.
\end{aligned}
}
\label{eq:surrogate_io}
\end{equation}
\RevisionBlue{The Tabor parameter $\mu$ is retained as an essential conditioning variable because it describes the adhesion regime independently of the prescribed kinematics. If $\mu$ were omitted, the same time--indentation history at different Tabor parameters could produce different force trajectories, so the reduced input would not define a unique target over the full dataset.}
Here, $\mathbf{X}$ contains all time-varying quantities provided to the network, where $d$ is the number of sequential features provided, also known as channel number. Furthermore, $\mu$ is supplied separately as a static scalar due to its nature. The loading rate, unloading rate, dwell duration, and maximum indentation are therefore represented through the sampled time-indentation history in $\mathbf{X}$, not as separate static inputs. This design choice is intended to support the model's potential for future generalization. We use two different types of feature definitions, where the purely data-driven representation uses:
\begin{equation}
    \mathbf{X}=\left[\mathbf{\widehat{t}, \widehat{\delta}} \right]\,,
\end{equation}
containing time vetor $\widehat{\mathbf{t}}$ and displacement vector $\widehat{\mathbf{\delta}}$ while the physics-guided representation has four channels as:
\begin{equation}
    \mathbf{X}=\left[\widehat{\mathbf{t}}, \widehat{\mathbf{\delta}},\mathbf{\dot{\widehat{\delta}}},\mathbf{e}_j  \right] \,,
\end{equation}
where $\mathbf{\dot{\widehat{\delta}}}$ is the velocity vector, and $\mathbf{e}_j$ is edge vector. The two-channel representation is a minimal data-driven description of the prescribed protocol. The four-channel representation augments it with quantities that make the loading-history structure more explicit. The velocity channel is computed causally through finite difference method (FDM) from the FMS-resampled indentation sequence as $\dot{\widehat{\delta}}_j=(\widehat{\delta}_j-\widehat{\delta}_{j-1})/(\widehat{t}_j-\widehat{t}_{j-1})$ for $j=2,\ldots,N$. The edge channel $e_j\in\{0,1\}$ is zero except where abrupt changes in the finite-difference velocity exceed the detection threshold, thereby highlighting transitions between loading, dwell, and unloading. \RevisionBlue{It is a deterministic, engineered encoding of the prescribed protocol, not an independently measured physical signal.} Unless otherwise stated, all experiments use the physics-guided four-channel representation; \ref{app:feature_comparison} compares it with the two-channel data-driven representation and shows that removing the velocity and edge channels produces a substantial increase in validation error.

\RevisionBlue{For reproducibility, the training script makes a shuffled, trajectory-level split using two calls to \texttt{train\_test\_split} with \texttt{random\_state=42}: the first reserves $20\%$ of the trajectories, and the second divides that subset equally into validation and test sets, giving an $80\%/10\%/10\%$ training/validation/test split. The symmetric logarithmic transform $s(x)=\operatorname{sgn}(x)\log_{10}(1+|x|)$ is applied twice to time and once to indentation, force, and the Tabor parameter. Velocity and edge channels are computed from the untransformed time--indentation histories. Separate channel-wise \texttt{StandardScaler} objects are fitted to the training trajectories only, sequence-channel statistics pool the training time steps, while the Tabor and force scalers use training samples and training force time steps, respectively. The fitted scalers are then applied unchanged to validation and test data. Thus, recreating the exact partitions requires the same processed files and row order, in addition to the stated split seed.}


\begin{figure}[htbp]
    \centering
    \includegraphics[width=0.92\linewidth]{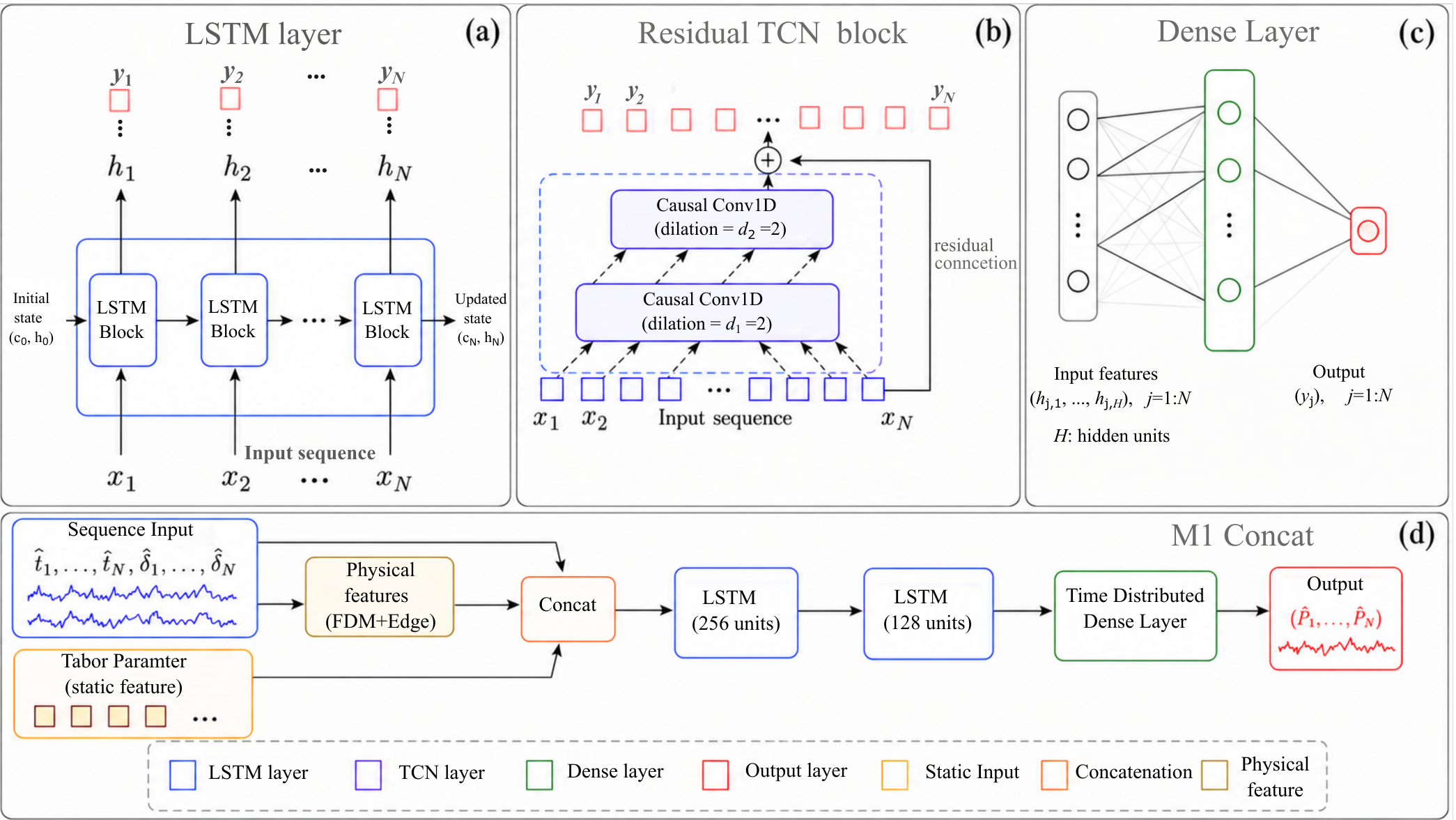}

    \caption{Schematic overview of the neural-network components and representative sequence-to-sequence architecture considered in this study: (a) LSTM module, (b) residual TCN block with causal convolutions, (c) dense layer, and (d) the M1-concat reference architecture (LSTM 256$\to$128 with concatenated Tabor conditioning and TimeDistributed Dense output layer), which achieves the best test-set performance among the eighteen compared models. The block labeled \textit{Physical features (FDM+Edge)} in panel~(d) denotes the two derived input channels appended to the base sequence: the finite-difference method (FDM) velocity $\dot{\widehat{\delta}}_j$ and the binary edge-detection (Edge) indicator $e_j$ that marks rapid rate transitions at the onset and end of the dwell phase (see Section~\ref{sec:methods} for definitions).}
    \label{fig:sequence_layers}
\end{figure}

\subsection{DL Models' architecture}\label{sec:s2s} 
As discussed in the previous sections, the considered viscoelastic material exhibits memory effects which means the adhesion force at a given step is not determined solely by the current displacement, but also by the preceding deformation history. In other words, the current state depends on previous states, which themselves carry information from earlier loading steps; this history-dependent behavior is what distinguishes the viscoelastic response from a purely elastic one. Similar requirements arise in other sequence-learning problems, such as speech recognition, audio processing, and machine translation, where the meaning of the current signal depends on its temporal context. This motivates the use of state-aware sequence models capable of preserving ordering and accumulating history. Figure~\ref{fig:sequence_layers} summarizes the three neural-network building blocks used for this purpose. Panel~(a) shows a Long Short-Term Memory (LSTM) layer \cite{hochreiter1997long}, in which the hidden cell states are updated while the sequence is traversed. This makes LSTM layers natural candidates for approximating the fading-memory character of the Boltzmann convolution in Eq.~(\ref{eq:uz}). Panel~(b) shows a residual Temporal Convolutional Network (TCN) block \cite{bai2018empirical}, where causal one-dimensional convolutions with dilation extract local and progressively wider temporal patterns without using a recurrent state. This provides a complementary way of encoding the loading history and allows the comparison between recurrent and convolutional memory representations. Panel~(c) shows the dense readout used in a time-distributed form: the same affine map is applied to the latent features at each measurement step, producing one scalar force prediction while preserving the sequence length. The mathematical definitions of the LSTM cell, residual TCN block, and TimeDistributed Dense readout are given in \ref{app:NN_components}. All neural-network models were implemented in Python using TensorFlow/Keras 2.18.0.

These building blocks were then combined into six families of sequence models, denoted M1--M6. M1 is the LSTM architecture, with two recurrent layers followed by a TimeDistributed Dense output layer. M2 adds layer normalization to the LSTM structure. M3 first applies one-dimensional convolutional layers to extract local temporal features and then processes the resulting sequence with an LSTM layer. M4 is a pure TCN model composed of four residual dilated causal blocks. M5 uses a transformer-encoder architecture as an attention-based alternative to recurrent and convolutional sequence encoders. M6 is a residual LSTM in which a skip connection combines two recurrent paths before the final recurrent layer.

For each of the six families, the static Tabor parameter $\mu$ was supplied through one of three conditioning mechanisms. In \textit{concatenation} conditioning, $\mu$ is repeated over the $N$ measurement steps and appended to the sequence features before the first temporal layer. In \textit{FiLM} conditioning, a small dense sub-network driven by $\mu$ produces feature-wise affine scale-and-shift coefficients applied to a hidden representation. In \textit{gated} conditioning, $\mu$ parameterizes a softmax gate that forms a convex mixture of two compatible temporal branches. These three alternatives are shown by the dashed conditioning paths in Figure~\ref{fig:modelarch} and are defined mathematically in \ref{app:tabor_conditioning}. Combining the six families with the three Tabor-conditioning mechanisms gives 18 sequence-to-sequence models. All eighteen models use the physics-guided four-channel input representation described in Section~\ref{sec:dm}. Figure~\ref{fig:sequence_layers}(d) illustrates the M1 architecture with concatenated Tabor conditioning. It is notable that this model is later identified as the best-performing configuration in \ref{sec:results} and is adopted as the reference surrogate. The complete set of M1--M6 architectures is schematically represented in \ref{sec:appendix_arch}, Figure~\ref{fig:modelarch}.

\subsection{Training and hyperparameters}

\begin{figure}[htbp]
    \centering
        \centering
        \includegraphics[width=0.48\linewidth]{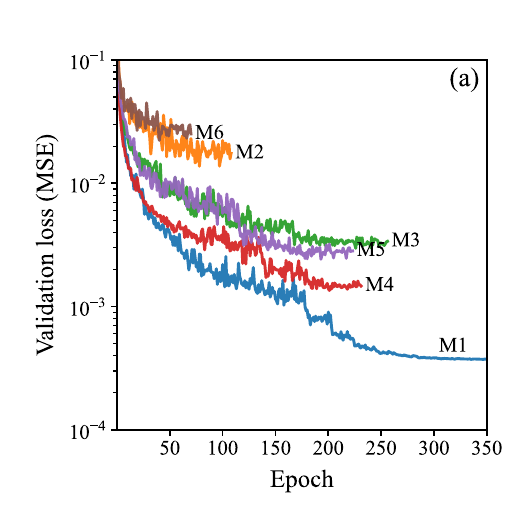}
        \centering
        \includegraphics[width=0.48\linewidth]{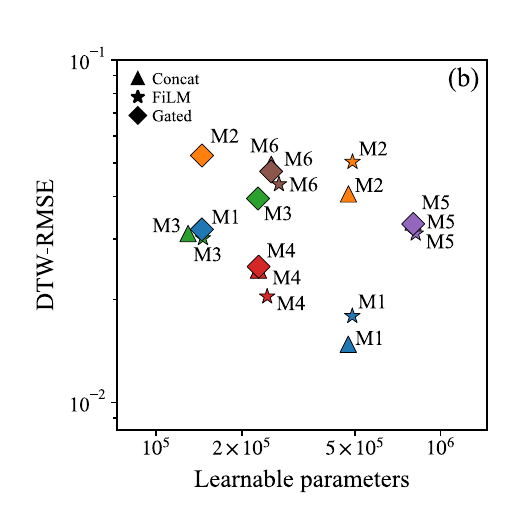}
    \vspace{0.25em}
        \includegraphics[width=0.48\linewidth]{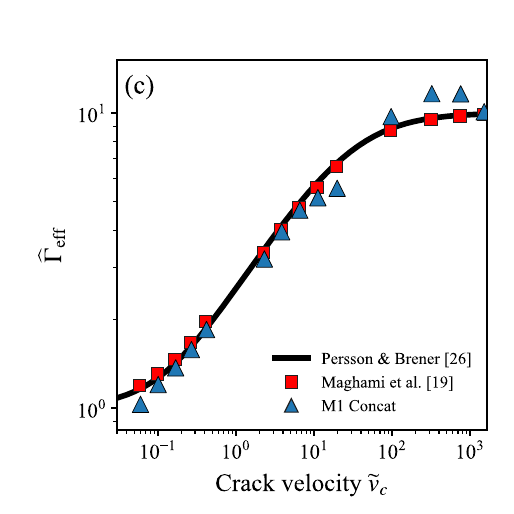}
        \hfill
        \includegraphics[width=0.48\linewidth]{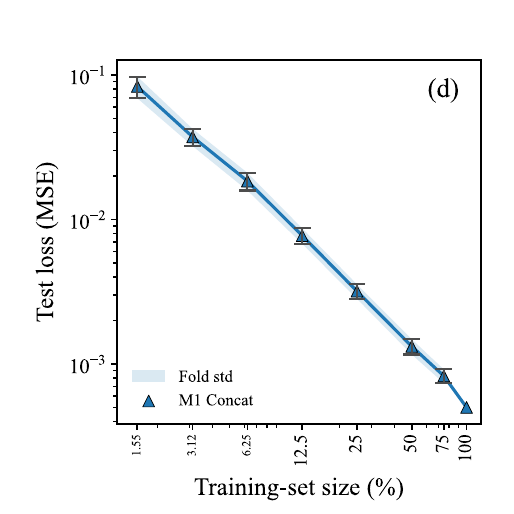}
    \caption{Model selection, physical validation, and data-efficiency diagnostics for the M1-concat surrogate. \textbf{(a)} Validation-loss learning curves for the concatenation-conditioned models. \textbf{(b)} Model size, measured by the number of learnable parameters, versus shape-sensitive trajectory error measured by DTW-RMSE. Marker shapes denote the conditioning mechanism: triangle = concatenation, star = FiLM, and diamond = gated conditioning; colors identify the six neural-network architectures M1--M6. {\textbf{(c)} Normalized effective surface energy $\widehat{\Gamma}_{\mathrm{eff}}=\Delta\gamma_{\mathrm{eff}}/\Delta\gamma_0$ versus normalized crack velocity $\widetilde{v}_c$. Red squares show the portion of the 19-point SLS numerical series from Figure~6 in Ref. \cite{maghami2024bulk} that lies within the displayed crack-velocity range, the black solid curve is the corresponding Persson \& Brener theory \cite{persson2005crack} prediction, and the fourteen blue triangles are M1-concat predictions evaluated at $\mu=3.24$ and at the selected deposited SLS unloading rates from the inset of Figure~7 in \cite{maghami2024bulk}, then plotted at their directly paired inset crack velocities \cite{maghami2024bulk,papangelo2024jmpsdata,persson2005crack}.} \textbf{(d)} Data-efficiency test for the selected M1-concat architecture: the 100\% point is the full-training reference model, whereas the reduced training fractions report the mean held-out test MSE over five nested training-subset folds; the shaded band and error bars denote one sample standard deviation across folds. Validation and test protocols are held fixed, so the trend isolates the effect of reducing the number of available BEM trajectories. Further learning-curve and feature-representation diagnostics are reported in~\ref{app:feature_comparison}.}
    \label{fig:tier1d_pareto}
\end{figure}

\RevisionBlue{The 18 final architecture--conditioning runs can be reproduced using TensorFlow/Keras 2.18.0 and the training data described above. Before the sweep, the script sets the Python, NumPy, and TensorFlow random seeds to 1337 and requests deterministic TensorFlow operations where supported; the same model and conditioning loop order should be retained because the random-number stream advances across runs. Within each conditioning group (concatenation, FiLM, then gated), the six model families are trained in order. Each model is optimized with Adam \cite{kingma2014adam}, an initial learning rate of $10^{-4}$, standardized-force MSE loss, and a batch size of 8, for at most 2000 epochs. Keras's default shuffling of training trajectories is retained. The learning-rate scheduler monitors validation loss and halves the rate after 20 non-improving epochs, down to $10^{-7}$. Early stopping monitors the same validation loss, stops after 30 non-improving epochs, and restores the weights from the best validation epoch. The concatenation-conditioned learning curves are shown in Figure~\ref{fig:tier1d_pareto}(a). Curves for the other 12 model--conditioning combinations are provided in Figure~\ref{fig:tier1d_learning_curves_appendix} in Section~\ref{sec:appendix_lc}. The released model weights and fitted scalers as well as per-epoch histories are stored in the repository of the paper. The training process was consucted on a system with a 12th Gen Intel Core i9-12900K processor, 16 cores, and 96 GB RAM. Validation data control the learning-rate schedule, early stopping, and restoration of the best weights. Hence, they are not used for gradient updates. The final 10\% test subset is likewise excluded from fitting and these callbacks, and is scored after each run.}


\section{Results}\label{sec:results}

All results reported in this section refer to the test set unless stated otherwise. We first use the model-comparison study to identify a reference sequence model for the remainder of the analysis, and then assess whether that model reproduces the canonical adhesion regimes and physically interpretable quantities relevant to detachment.

\subsection{Model-family selection}
Here, model comparison was based on global waveform metrics as well as MSE and mean absolute error (MAE). When comparing two curves or time-dependent signals, particularly in mechanical and engineering applications, MSE and MAE alone may be insufficient, since the global shape, phase, and qualitative evolution of the response can be more relevant than strictly point-to-point discrepancies. Therefore, in addition to these classical error measures, we considered dynamic time warping-root MSE (DTW-RMSE) \cite{giorgino2009computing}. DTW-RMSE combines dynamic time warping with root mean squared error by first aligning two curves through an optimal warping path and then computing the RMSE over the aligned samples, making it suitable for signals with similar shapes but local temporal shifts or distortions \cite{giorgino2009computing, cuturi2017soft}.

Figure~\ref{fig:tier1d_pareto} compares all eighteen models provided in \ref{sec:appendix_arch} using validation-learning behavior, shape-sensitive trajectory error, model size, and a test against analytical solution. The model denoted M1 with concatenated Tabor conditioning (blue triangle in Figure \ref{fig:tier1d_pareto}b) provides the best overall balance among the compared candidates when accuracy and learnable-parameter count are considered together. In particular, it attains a held-out MSE of $5.0 \times 10^{-4}$, an MAE of $9.2 \times 10^{-3}$, a DTW-RMSE of $1.48 \times 10^{-2}$, and uses 473,729 learnable parameters. Within this comparison, the concatenation conditioning strategy is therefore sufficient to outperform the FiLM and gated alternatives, and M1-concat is adopted as the reference model in the remaining results. The remaining validation learning curves are provided in \ref{sec:appendix_lc}. \textcolor{black}{The repeated-seed comparison in \ref{sec:appendix_lc} supports this choice. \ref{app:fms_resolution_transfer} complements the representative resolution-transfer check with separately trained models evaluated against the original BEM trajectories.}

The data-fraction experiment in Figure~\ref{fig:tier1d_pareto}(d) was introduced to distinguish architectural adequacy from simple data abundance and to estimate how rapidly predictive accuracy deteriorates when fewer BEM trajectories are available. The point at 100\% corresponds to the selected reference M1-concat model trained on the full training partition, whereas each reduced-fraction point is the mean over five retrainings on nested subsets of the same training partition. The validation and test partitions were kept identical to those used for the reference model, so changes in error can be attributed to the amount and coverage of available training information. The shaded band and error bars report the fold-to-fold standard deviation. The error increases progressively as the training fraction is reduced. \ref{app:feature_comparison} reports the associated validation histories for one fold.

The trends in Figure~\ref{fig:tier1d_pareto}, together with the validation curves in \ref{sec:appendix_lc} (Figure~\ref{fig:tier1d_learning_curves_appendix}), reveal two effects. First, the conditioning mechanism matters because $\mu$ is a continuous adhesion-regime variable, controlling the DMT--JKR transition and cohesive-zone scale, while the network must also encode the viscoelastic loading history. The comparison does not indicate a universal superiority of concatenation conditioning. Based on the held-out MSE, concatenation performs best within M1 and M2, FiLM within M3 to M5, and gated conditioning within M6 (see Figure \ref{fig:tabor_conditioning_mse_mae}). Thus, M1-concat is the best individual configuration, while the relative effectiveness of each conditioning mechanism depends on the temporal family and injection point. In M1-concat, $\mu$ enters the LSTM at every measurement step, so the same recurrent gates learn both the fading-memory response associated with the Boltzmann convolution and the adhesion regime. Second, the family comparison shows that architectural simplicity and the internal memory are advantageous. The LSTM-based architecture (M1) gives the best selected model, while the TCN family (M4) is the strongest alternative, consistent with the fact that causal dilated convolutions can represent finite loading-history windows. \textcolor{black}{One possible explanation for the observed advantage is that the LSTM cell state provides an input-dependent fading-memory pathway, whereas the TCN has a fixed, finite receptive field. In the implemented M4 model, this field spans a portion of the 120-step sequence, so late-step predictions cannot directly use the complete earlier loading--dwell history (\ref{sec:appendix_arch}). The LSTM can, in principle, carry a compressed and gated representation of the full preceding sequence, retaining or attenuating past information as the protocol evolves. This may better match the cumulative but fading history dependence represented by the Boltzmann convolution, particularly across different rates and dwell durations.} As shown in Figure \ref{fig:tier1d_pareto}(b), comparing M1 Concat with the M5 family of models, which contains a larger number of learnable parameters, it can be inferred that the number of parameters alone is not the primary factor determining performance. Rather, the way these parameters interact and are utilized within the model architecture plays a more significant role.

\subsection{Verification in the fully relaxed short-range adhesion limit}\label{sec:relaxed_short_range_verification}

For the general loading--dwell--unloading histories considered in this work, no closed-form analytical solution is available for the complete force trajectory. An analytical reference for detachment is available only in the fully relaxed, short-range adhesion limit and with sufficiently large preload, in which the contact is JKR-like and its receding edge can be treated as a viscoelastic interfacial crack. We therefore use this restricted regime as a controlled physical verification of M1-concat against the numerical study of Maghami et al.\ \cite{maghami2024bulk} and the Persson--Brener (PB) crack-propagation theory \cite{persson2005crack}, rather than as a validation over the entire protocol space.

\RevisionBlue{We set $\mu=3.24$ to match the SLS parameter used in the study of Maghami et al. \cite{maghami2024bulk}, allowing comparison with its numerical results and the corresponding PB prediction. Since our training range ends at $\mu=3.2$, this is a controlled 1.25\% extrapolation in Tabor parameter. The fourteen reference SLS BEM cases are taken from the separate dataset accompanying that study \cite{papangelo2024jmpsdata}. They were not included in our training, validation, or test partitions and were not used for model or hyperparameter selection. We use this external dataset and the PB solution only for a post-training physical-consistency check.}

The common comparison quantity is the rate-dependent effective surface energy $\Delta\gamma_{\mathrm{eff}}(\widetilde{v}_c)$ that is the macroscopic energy per unit newly separated area required to advance the contact edge. Here $v_c$ denotes the contact-edge crack velocity and $\widetilde{v}_c=v_c\tau_r/l_0$ its normalization, with $\tau_r$ and $l_0=E_0^*\Delta\gamma_0/[\pi(\alpha\sigma_0)^2]$ denoting the characteristic material time and PB length scale \cite{maghami2024bulk}, respectively. The parameter $\alpha\approx\pi/9$ is an empirical coefficient reported by \cite{maghami2024bulk} to relate the LJ maximum tensile stress $\sigma_0$ with the critical stress $\sigma_c$ introduce in PB theory \cite{persson2005crack}. The subscript $c$ distinguishes crack velocity from the prescribed loading and unloading velocities $v_L$ and $v_U$. Within linear viscoelastic fracture mechanics, $\Delta\gamma_{\mathrm{eff}}$ combines the thermodynamic work of adhesion $\Delta\gamma_0$ with the bulk viscoelastic dissipation accompanying crack propagation and is therefore a crack-velocity-dependent fracture quantity, rather than an additional constant surface property. We use the normalized form $\widehat{\Gamma}_{\mathrm{eff}}=\Delta\gamma_{\mathrm{eff}}/\Delta\gamma_0$. In the fully relaxed short-range limit, and provided that the initial contact is sufficiently large to suppress finite-size effects, this quantity is related to the pull-off amplification by $\widehat{\Gamma}_{\mathrm{eff}}\simeq |P_{\mathrm{po}}|/P_{\mathrm{JKR}}$, where $P_{\mathrm{JKR}}=1.5\pi R\Delta\gamma_0$. PB theory supplies $\widehat{\Gamma}_{\mathrm{eff}}$ directly as a function of $\widetilde{v}_c$, which is used in Figure~\ref{fig:tier1d_pareto}(c) \cite{maghami2024bulk}.

To approach the assumptions of this analytical limit with the present surrogate, all protocol quantities other than unloading velocity are fixed at $\mu=3.24$, $\widehat{\delta}_l=80$, $\widehat{v}_L=1.3895$, and $\widehat{t}_D=3.0$. The high Tabor parameter and large peak indentation promote short-range, preload-independent JKR-like detachment, while the long loading duration followed by the dwell brings the substrate close to its relaxed state before retraction. The unloading velocities are not chosen from an auxiliary grid, and they are exactly the fourteen selected SLS retraction rates associated with the numerical results of Maghami et al.\ \cite{maghami2024bulk}. Because PB theory is expressed in crack-velocity space, each M1-concat prediction is assigned the crack velocity paired directly with that retraction rate in the inset of Figure~7 of Ref.~\cite{maghami2024bulk}. The details are documented in ~\ref{sec:appendix_fig2}.

M1-concat predicts the complete force trajectory for each of these fourteen protocols, and $|P_{\mathrm{po}}|$ is extracted a posteriori as the maximum tensile-force magnitude during unloading. Figure~\ref{fig:tier1d_pareto}(c) compares the resulting $\widehat{\Gamma}_{\mathrm{eff}}$ values (blue triangles) with both the SLS BEM results (red squares) of Ref \cite{maghami2024bulk} and the corresponding PB solution (solid black line). The surrogate follows the reference trend closely at low and intermediate crack velocities. The larger deviations at the highest crack velocities are consistent with the increasingly localized pull-off event, for which the predicted maximum is more sensitive to the finite resolution of the fixed-measurement-step representation. Note that neither the PB nor the samples in Ref \cite{maghami2024bulk} were used in the training process of M1-Concat.  We also note that the M1 Concat is trained to predict the whole trajectory on the whole training domain, while this test was to validate an abstract quantity on the specific relaxed short-range adhesion regime.

\RevisionBlue{We use PB as an independent analytical benchmark to test whether the time-resolved surrogate also recovers this established limiting behavior \cite{maghami2024bulk}. Agreement with the BEM results assesses numerical emulation, whereas agreement with PB is a regime-specific theory-consistency check.}

\subsection{Canonical adhesion regimes}
\begin{figure}[htbp]
    \centering
\includegraphics[width=0.49\linewidth]{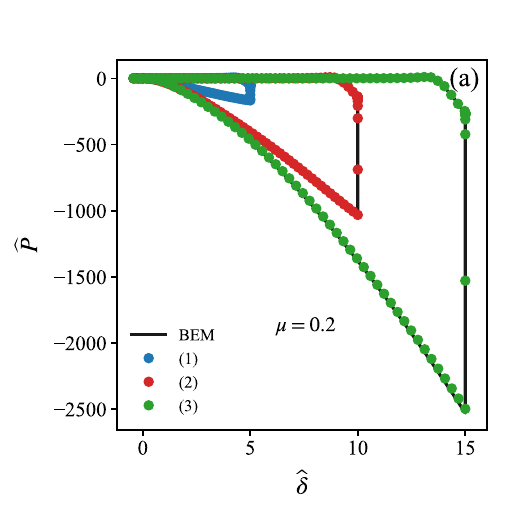}
    \hfill
\includegraphics[width=0.49\linewidth]{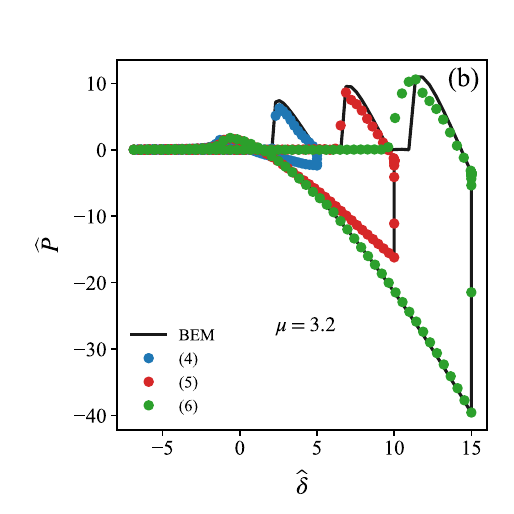}\\
    \vspace{0.5em}
\includegraphics[width=0.49\linewidth]{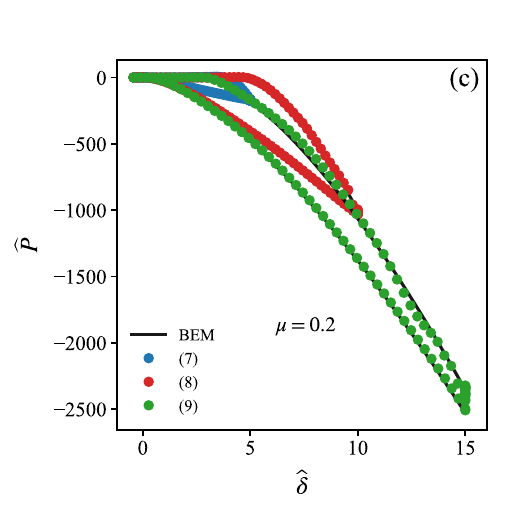}
    \hfill
\includegraphics[width=0.49\linewidth]{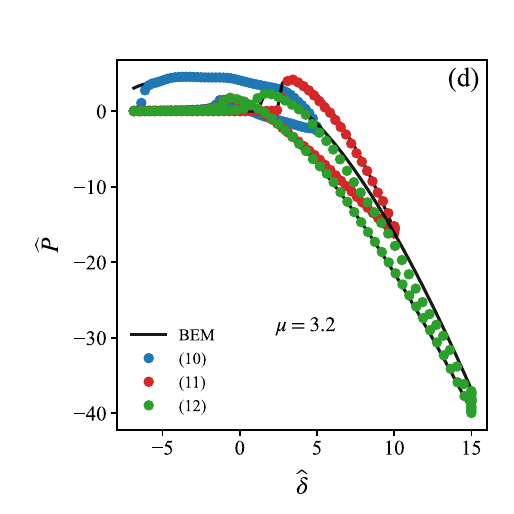}

\caption{Force--indentation response of the M1-concat surrogate for the twelve representative unseen protocols listed in Table~\protect\ref{tab:fig9_protocol_parameters}. Solid black curves show the BEM reference simulations, while colored markers show the surrogate predictions; the panel legends identify samples (1) to (12), and the in-panel labels report the corresponding Tabor parameter. Panels (a) and (c) show low-Tabor cases with $\mu=0.2$, whereas panels (b) and (d) show high-Tabor cases with $\mu=3.2$. Panels (a,b) correspond to longer-dwell protocols, while panels (c,d) correspond to short-dwell protocols with increasing loading rate, unloading rate, and peak indentation from the first to the third sample in each panel. Full loading parameters, BEM runtimes, and M1 Concat inference times are reported in Table~\protect\ref{tab:fig9_protocol_parameters}.}
    \label{fig:canonical_behaviors}
\end{figure}
The predictions of the M1-concat model are shown in Figure~\ref{fig:canonical_behaviors} (for unseen data) to allow a direct comparison between the numerical BEM predictions (solid black line) and the M1 Concat ones (markers) for 12 different combinations of input parameters. It is worthwhile to recall that although the loading protocol is a function of 6 input parameters $\{\widehat{\delta}_0,\widehat{v}_L,\widehat{v}_U,\widehat{t}_D,\widehat{\delta}_l,\mu\}$ the surrogate model sees the displacement history plus the Tabor parameter. In Figure~\ref{fig:canonical_behaviors} we considered: (a) low Tabor parameter in combination with relatively long dwell time, (b) high Tabor parameter in combination with relatively long dwell time, (c) low Tabor parameter combined with very short dwell time, (d) high Tabor parameter combined with short dwell time. The parameter combinations are labeled with a number from 1 to 12, and the loading parameters are reported in Table \ref{tab:fig9_protocol_parameters}. Hence, Figure~\ref{fig:canonical_behaviors} shows a combination of paradigmatic loading conditions in viscoelastic contacts, considering unloading paths that start when the substrate is relaxed or immediately after a rapid loading sequence in combination with different adhesion regimes, from short to long-range. Across all the tests, the surrogate reproduces the overall loop shape, the location of the pull-off event, and the inherent differences between glassy-like and rubbery-like responses. The agreement is particularly relevant in panels (c) and (d), where the short dwell and high rates preserve the memory of the preceding loading history and therefore provide the strongest test of the sequence model. Figure~\ref{fig:canonical_behaviors} demonstrates that the M1-concat provides consistent predictions for all the combinations tested and not only along the main loading-unloading branches but also when single points of particular interest are extracted (e.g., the pull-off force or the maximum preload). Notice also that Ref. \cite{maghami2025pull} trained a machine learning model to predict the pull-off force when a viscoelastic substrate was unloaded from a fully relaxed condition. Here, this limitation is overcome as pull-off predictions show strong agreement for all the protocols tested.
The remaining visible errors are concentrated near the sharp adhesive-instability and snap-off portions of the unloading branch, most clearly in the high-Tabor cases of panel~(d and b), where a small shift in the transition position produces a large local force discrepancy.
  
\begin{table}[h]
    \centering
   \caption{Loading-protocol parameters and computational timings for the numbered representative samples shown in Figure~\protect\ref{fig:canonical_behaviors}.}
    \label{tab:fig9_protocol_parameters}
    \scriptsize
    \setlength{\tabcolsep}{2pt}
    \begin{threeparttable}
    \begin{tabularx}{\linewidth}{@{}cCCCCCCC@{}}
        \toprule
        Sample & {\shortstack{Loading\\rate $\widehat{v}_L$}} & \shortstack{Dwell time\\$\widehat{t}_D$} & {\shortstack{Unloading\\rate $\widehat{v}_U$}} & \shortstack{Tabor\\parameter $\mu$} & \shortstack{Peak indentation\\$\widehat{\delta}_l$} & \shortstack{BEM time\\(s)} & \shortstack{M1 Concat time\\(s)} \\
        \midrule
        (1) & $10$ & $1$ & $200$ & $0.2$ & $5$ & $658$ & $0.12$ \\
        (2) & $100$ & $2$ & $200$ & $0.2$ & $10$ & $456$ & $0.21$ \\
        (3) & $1000$ & $3$ & $200$ & $0.2$ & $15$ & $488$ & $0.26$ \\
        (4) & $10$ & $1$ & $1000$ & $3.2$ & $5$ & $6585$ & $0.17$ \\
        (5) & $100$ & $2$ & $1000$ & $3.2$ & $10$ & $5764$ & $0.16$ \\
        (6) & $1000$ & $3$ & $1000$ & $3.2$ & $15$ & $1214$ & $0.16$ \\
        (7) & $10$ & $1.0\times10^{-3}$ & $10$ & $0.2$ & $5$ & $333$ & $0.15$ \\
        (8) & $100$ & $4.5\times10^{-3}$ & $100$ & $0.2$ & $10$ & $364$ & $0.15$ \\
        (9) & $1000$ & $8.0\times10^{-3}$ & $1000$ & $0.2$ & $15$ & $424.4$ & $0.23$ \\
        (10) & $10$ & $1.0\times10^{-3}$ & $10$ & $3.2$ & $5$ & $407.5$ & $0.23$ \\
        (11) & $100$ & $4.5\times10^{-3}$ & $100$ & $3.2$ & $10$ & $258.2$ & $0.13$ \\
        (12) & $1000$ & $8.0\times10^{-3}$ & $1000$ & $3.2$ & $15$ & $254.2$ & $0.14$ \\
        \bottomrule
    \end{tabularx}
    \begin{tablenotes}[flushleft]
        \item Reported BEM times correspond to the last successful time-discretization attempt. Although warm-up discretization and adaptive time steps were used in BEM, convergence may require additional trial-and-error attempts, which are not included in the computational-cost comparison.
    \end{tablenotes}
    \end{threeparttable}
\end{table}

Table \ref{tab:fig9_protocol_parameters} reports the loading properties, the Tabor parameter ($\mu$), the simulation time, and the network inference time, showing that the M1 Concat model provides predictions about three orders of magnitude faster than a classical BEM numerical approach, regardless of the input parameters. Notice that this may play a crucial role in integrating real-time estimation of the contact state in practical engineering applications (grasping, crawling, manipulation) or in multi-body multi-degrees of freedom systems.

\subsection{Physically interpretable error statistics}

Table ~\ref{tab:quartile_metrics} stratifies prediction accuracy by sorting the test set 1245 evaluated samples into four equal groups ranked by per-sample force MSE, from Q1 (lowest-error quartile) to Q4 (highest-error quartile). Notice that the force MSE metric is computed over the whole trajectory; this may differ from the point-wise error on a single quantity, such as the pull-off force, the time at which pull-off happens, or the enclosed hysteresis loop. This stratification identifies which error regimes drive the overall figures and reveals whether physically central quantities such as pull-off force and enclosed hysteresis-loop degrade in concert with the global force-sequence error. 

For the $i$-th trajectory, the pull-off force $\widehat{P}^{(i)}_{\mathrm{po}}$ is defined as the maximum tensile force reached during the unloading phase, and the pull-off time $\widehat{t}^{(i)}_{\mathrm{po}}$ is the corresponding time instant. The enclosed hysteresis-loop area is evaluated in the normalized force--indentation plane as $\widehat{\mathcal{A}}^{(i)}_{\mathrm{hys}}=\oint \widehat{P}^{(i)}\,d\widehat{\delta}^{(i)}$. The errors reported for pull-off force, pull-off time, and hysteresis area are relative errors, computed as $100|\widehat{y}^{(i)}_{\mathrm{M1}}-{y}^{(i)}_{\mathrm{BEM}}|/|{y}^{(i)}_{\mathrm{BEM}}|$ where $\mathrm{M1}$ is the predicted values by M1 Concat and ${y}^{(i)}_{\mathrm{BEM}}$ is the BEM reference results. The values in Table~\ref{tab:quartile_metrics} are the mean and standard deviation of these errors within each force-MSE quartile.

\begin{table}[h]
    \centering
    \caption{Prediction errors by test-set quartile. Samples are ranked by per-sample force MSE and divided into four equal groups, from Q1 (lowest-error quartile) to Q4 (highest-error quartile). Entries are reported as mean $\pm$ standard deviation.} 
    \label{tab:quartile_metrics}
    \begin{threeparttable}
        \small
        \setlength{\tabcolsep}{4pt}
        \begin{tabularx}{\linewidth}{@{}L{2.1cm}CCCC@{}}
            \toprule
            Quartile
                & \shortstack{Force MSE\\($\times10^{-3}$)}
                & \shortstack{Pull-off\\Error (\%)}
                & \shortstack{Hysteresis\\Error (\%)}
                & \shortstack{Pull-off Time\\Error (\%)} \\
            \midrule
            Q1 & $0.0853\pm0.0229$ & $2.4\pm2.5$ & $1.7\pm2.7$ & $0.6\pm1.5$ \\
            Q2 & $0.1587\pm0.0246$ & $3.2\pm3.7$ & $2.0\pm4.7$ & $0.6\pm1.4$ \\
            Q3 & $0.2939\pm0.0630$ & $4.0\pm5.1$ & $2.3\pm4.0$ & $0.6\pm2.0$ \\
            Q4 & $1.5134\pm1.6845$ & $6.1\pm8.9$ & $3.1\pm6.0$ & $0.4\pm1.1$ \\
            \bottomrule
        \end{tabularx}
        \begin{tablenotes}[flushleft]
            \footnotesize
           \item Pull-off value, pull-off time, and hysteresis entries are relative errors in percent, while force MSE is not a relative error. The force-MSE column reports $10^{3}\mathrm{MSE}$. Mean $\pm$ standard deviation is computed across the per-sample errors within each force-MSE quartile.
        \end{tablenotes}
    \end{threeparttable}
\end{table}

Table~\ref{tab:quartile_metrics} summarizes how the prediction errors evolve from the lowest- to the highest-error trajectories in the test set. In the highest-error quartile, the mean pull-off error remains at $6.1\%$, while the mean hysteresis error is $3.1\%$ and the mean pull-off-time error remains below $1\%$. This indicates that even in the Q4 quartile, deviations from the numerical simulations are concentrated mainly in the detailed waveform reconstruction, whereas the physically central detachment quantities remain comparatively well captured over most of the test set. \RevisionBlue{A low global MSE does not by itself guarantee accurate prediction near snap-off, where the rapid force change makes local force error sensitive to small shifts in event timing. In the highest-error quartile, mean pull-off-force error is $6.1\%$ despite mean pull-off-time error below $1\%$ (Table~\ref{tab:quartile_metrics}), showing that event timing and force amplitude can have different accuracy. Figure~\ref{fig:fms_resolution_transfer}(b) complements them by showing pull-off-force and hysteresis errors stratified by unloading-rate quartile.}

\begin{figure}[htbp]
    \centering
    \includegraphics[width=0.49\linewidth]{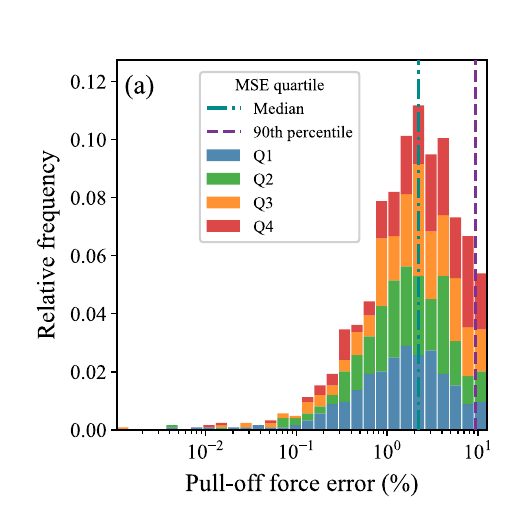}
    \hfill
    \includegraphics[width=0.49\linewidth]{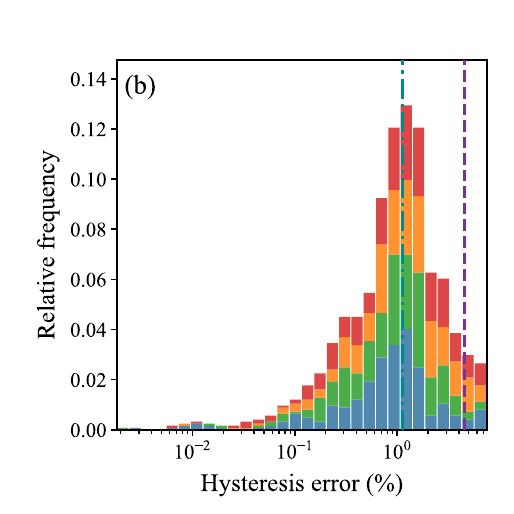}\\
    \vspace{0.5em}
    \includegraphics[width=0.49\linewidth]{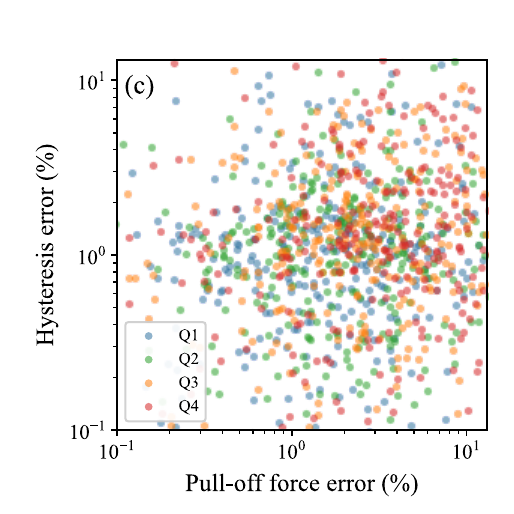}
    \hfill
    \includegraphics[width=0.49\linewidth]{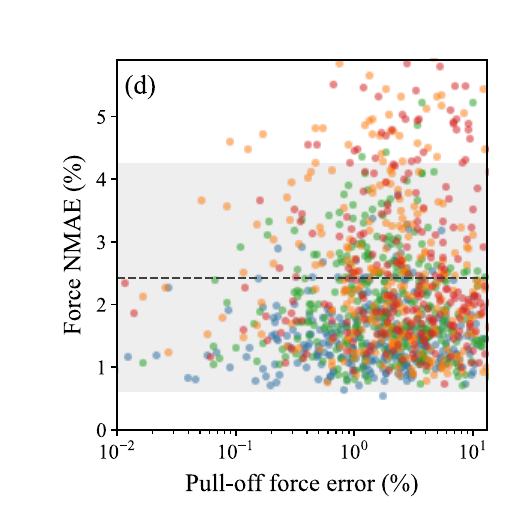}
\caption{Distribution and cross-correlation of physically interpretable prediction errors over the held-out test set, stratified by force-MSE quartile. Panels~(a) and (b) show relative-frequency histograms of pull-off force error and hysteresis error, respectively, using logarithmically spaced bins; both quantities use the same relative-error definitions as Table~\ref{tab:quartile_metrics}. Histogram bars are color-coded by force-MSE quartile (Q1: lowest-error quartile, through Q4: highest-error quartile). Vertical dash-dotted and dashed lines indicate the empirical median and 90th percentile, respectively. Panel~(c) compares pull-off force error with hysteresis error on logarithmic axes. Panel~(d) compares pull-off force error with the global force normalized mean absolute error, $\mathrm{NMAE}=100\sum_j|\widehat{P}^{\mathrm{M1}}_j-\widehat{P}^{\mathrm{BEM}}_j|/\sum_j|\widehat{P}^{\mathrm{BEM}}_j|$.}
    \label{fig:error_distribution}
\end{figure}

A more detailed analysis is shown in Figure~\ref{fig:error_distribution} where the relative-frequency distributions of pull-off force error (panel (a)) and hysteresis error (panel (b)) are shown. The colors within the bars in Figure~\ref{fig:error_distribution} refer to the four different MSE quartiles. Notice that both distributions are right-skewed, with most samples concentrated at low error values where one finds samples belonging to all the quartiles from Q1 to Q4, confirming that pull-off and hysteresis are generally predicted accurately even for test cases belonging to the Q4 quartile. From the error distribution, over the whole test dataset, the median error for the pull-off force is $\approx2.2\%$ while for the hysteresis it reads $\approx1.1\%$. Figure~\ref{fig:error_distribution} suggests the most severe inaccuracies remain confined to a relatively small subset of the test trajectories. Panels~\ref{fig:error_distribution}(c,d) further clarify that the different error measures probe distinct aspects of the predicted trajectory. Panel~(c) compares the relative pull-off force error with the relative hysteresis-loop error for each held-out trajectory and shows no simple one-to-one correspondence between them, which means trajectories with small pull-off error may still exhibit a non-negligible hysteresis error, and conversely. This could be attributed to the fact that the pull-off force is a local detachment quantity whereas the hysteresis area integrates the full loading--unloading loop. Panel~(d) relates the pull-off force error to the global force NMAE, with colors indicating the force-MSE quartiles used in Table~\ref{tab:quartile_metrics}. As expected, the quartiles are more clearly ordered along the force-NMAE axis, since both NMAE and MSE quantify global point-wise waveform accuracy. However, the pull-off errors remain broadly distributed within each NMAE range, indicating that global waveform metrics alone do not uniquely determine the accuracy of physically relevant scalar quantities.


\section{Discussion}\label{sec:discussion}
Within the viscoelastic contact setting and parameter ranges considered here, the results show that the full adhesive force trajectory can be predicted with low and controlled error and identify the most influential modeling choices among those examined. First, the model comparison shows that performance gains did not require the most elaborate architecture. The baseline LSTM with concatenated Tabor conditioning (M1-concat) provided the best overall compromise between waveform fidelity and physically meaningful scalar errors, which indicates that most of the problem difficulty lies in representing the loading history and the adhesion-regime parameter consistently rather than in increasing architectural complexity.

Second, the best-performing errors are not uniformly distributed across the test set. The canonical trajectories and quartile analysis show that the network reproduces smooth loading, dwell, and pull-off behavior reliably for most histories, while the largest deviations cluster in trajectories with sharper transitions and more localized force variations. . Such errors are physically plausible because abrupt changes in unloading rate and behavior near pull-off can increase sensitivity to constitutive relaxation and adhesion-range effects. The concentration of error in these cases suggests that the remaining limitations primarily concern histories with rapid transitions and localized force variations not captured properly with the defined FMS, rather than the learned mapping as a whole.

The model is not presented as a replacement for contact mechanics, nor as evidence that constitutive structure can be ignored. Rather, the network acts as a surrogate for a well-defined class of viscoelastic contact simulations parameterized by loading history and adhesion regime. 

The central objective of this work is therefore not to propose the most accurate possible numerical description of adhesive viscoelastic contact, but to provide a route to very rapid predictions with an accuracy that can be controlled and improved. Direct numerical methods remain indispensable as reference tools, but their computational cost depends strongly on the physical regime and on the spatial and temporal discretization choices; in practice, obtaining reliable solutions may require trial-and-error refinement, which is unfavorable for real-time or repeated-query applications. The accuracy reported here should thus be interpreted as a first surrogate benchmark rather than a final limit for this problem. Larger and higher-quality datasets, a larger number of fixed measurement-step points, and a more systematic exploration of neural-network hyperparameters are expected to further improve predictive performance. In closed-loop applications such as design and optimization, the surrogate can therefore be used for rapid exploration, while the final selected configuration should still be rechecked with high-fidelity simulation at the last designing stage.

\RevisionBlue{The computational advantage of the surrogate depends jointly on the required prediction accuracy and the number of subsequent queries. The amortized-cost analysis in ~\ref{app:data_costs} combines the observed mean test errors for different training fractions with the estimated costs of generating the corresponding database and fitting the model. Across the tested fractions, tighter mean-error targets are associated with larger offline investments and higher break-even query counts. Direct BEM can therefore have the lower total cost when only a few trajectories are required, particularly when the goal is to have limited high-fidelity level simulations. For sustained use involving many predictions within the validated domain, the DL+BEM framework can recover its initial cost and provide rapid responses, provided its errors satisfy the application requirements.}


\RevisionBlue{The scope also defines the main limitation. The trained surrogate represents a smooth, homogeneous half-space governed by a single-relaxation SLS with fixed modulus ratio $k=0.1$ and a Lennard--Jones traction--separation law. Its validation therefore demonstrates generalization only across the sampled loading histories and Tabor parameters within this model family. Transfer to broader relaxation spectra, including power-law rheologies, would require reference trajectories generated with those constitutive laws, followed by retraining and reevaluation of the network \cite{maghami2024bulk,bonfanti2020fractional}. Rough contacts would require roughness descriptors and representative training simulations \cite{papangelo2020numerical}, while layered, coated, or functionally graded media would require layer geometry and spatially varying material properties to be represented in the model inputs and training data; studies of layered functionally graded contact illustrate these additional mechanics \cite{revYaylaci2025}. A full experimental validation, including for prospective soft-robotic gripping applications, would require measuring force trajectories under controlled loading--dwell--unloading protocols, identifying material and adhesion parameters independently, and quantifying measurement uncertainty and model--experiment discrepancy. Until such extensions are trained and validated, applicability to these material classes, contact conditions, and experimental or vibration-driven settings remains unestablished. The difficult tail of the current error distribution also suggests that future datasets should improve coverage of fast unloading events and near-instability trajectories.}

\section{Summary and conclusions} \label{sec:conc}
This work examined whether deep sequence models can predict the full force trajectory of unsteady adhesive viscoelastic contact directly from the loading history and the Tabor parameter. Using BEM-generated simulations as reference data, we showed that a relatively simple LSTM-based architecture with concatenated Tabor conditioning is sufficient to reproduce the main loading, dwell, and unloading phases across a broad set of histories, while also achieving low errors in trajectory-level and physics-based summary metrics.

Two conclusions follow from the results. First, the main challenge is not architectural depth alone, but representation. Models that encode the loading history clearly and receive an explicit adhesion-regime descriptor perform best. Second, physics-guided feature 
\textcolor{black}{remains useful as an input-design strategy in the comparison examined here. With Tabor conditioning retained in both cases, augmenting the time--indentation input with velocity and edge-related features reduced validation error in the reported runs.} 

The manuscript therefore supports a focused claim. For adhesive viscoelastic contact, neural sequence surrogates can predict the complete time-resolved force response rather than only isolated events such as pull-off. The remaining errors are concentrated in a limited subset of sharply varying trajectories, which points to a concrete path forward of enriching the training set in those challenging regimes as well as increasing the FMS.

Specifically, the main findings are: (i) a BEM-generated dataset of \RevisionBlue{12,446} adhesive Hertzian contact trajectories covering loading rates from $10^{-1}$ to $10^{3}$, dwell times from $10^{-3}$ to 5, and Tabor parameters from 0.2 to 3 provides sufficient coverage for reliable sequence-model training; (ii) among the 18 compared models, the baseline LSTM with concatenated Tabor conditioning (M1-concat) achieves the best overall performance, confirming that architectural simplicity is preferable at this problem scale; \RevisionBlue{(iii) the combined engineered input representation, comprising causal velocity and a protocol-edge indicator, reduces validation MSE by approximately a factor of 7 relative to time--indentation alone. This comparison supports the utility of explicitly encoding the prescribed protocol, but does not attribute the gain to either channel individually or to new physical measurements;} and (iv) the model correctly reproduces low- and high-Tabor adhesion behaviors in both the relaxed and instantaneous limits, with median pull-off force error $\approx2.2\%$ and median hysteresis error $\approx1.1\%$ with accurate results provided also for the highest force MSE quartile.

Overall this work suggests that time-consuming numerical simulations of viscoelastic, soft, adhesive contacts may be effectively replaced by fast deep sequence networks, as the one proposed here. The advantage must not be reduced to the $\approx3$ orders of magnitude ``speed-up'' of the numerical analysis. The gain in terms of inference time may potentially open to unforeseen applications. For example, simulating the gripper-object contact ``while-grasping'' or the manipulation task in ``real-time'', may allow finer optimization in soft robotics and human-robot interactions, unfeasible with traditional numerical techniques. This work aimed to pave the way towards a family of surrogate contact models capable of rapid real-time prediction of the contact state in soft viscoelastic adhesive interfaces.  

Future work should address: augmenting the dataset in fast-unloading and near-instability regimes to improve the prediction accuracy in challenging scenarios; extending the framework to broader viscoelastic constitutive models; and validating the surrogate against experimental adhesion measurements for soft-contact applications.

\section*{Data and Code Availability}

\RevisionBlue{A browser-based demonstrator of the trained surrogate model, which enables users to define loading--dwell--unloading protocols and interactively inspect the predicted adhesive-force response, is available at
\begin{itemize}
  \item \url{https://alimaghamii.github.io/DL4Adhesion-demo}.
\end{itemize}
The data and code used for training, testing, and validation are available on Zenodo at}
\begin{itemize}
 \RevisionBlue{\item Zenodo: \url{https://zenodo.org/records/23086527}
  \item Github: \url{https://github.com/alimaghamii/DLViscoAdhesion}.}
\end{itemize}

\section*{\RevisionBlue{Declaration of generative AI and AI-assisted technologies in the manuscript preparation process}}
\RevisionBlue{During the preparation of this manuscript, the authors used ChatGPT to assist with language editing, including checking tone, spelling, and grammar. The authors reviewed and edited the suggestions as needed and take full responsibility for the final content of the published article.}

\section*{Acknowledgments}
This work was carried out as part of the Postdoctoral Fellowship of the first author, Ali Maghami, and received funding from the European Union's Horizon Europe research and innovation programme under the Marie Sk{\l}odowska-Curie Actions, Grant Agreement No. 101285102, ``Real-time Adhesion Trajectories \& Inverse Design via Physics-Enhanced Machine Learning'' (REAL-ADHERE). A.P. was supported by the European Union (ERC-2021-STG, ``Towards Future Interfaces With Tuneable Adhesion By Dynamic Excitation'' - SURFACE,
Project ID: 101039198, CUP: D95F22000430006). Views and opinions expressed are however those of the authors only and do not
necessarily reflect those of the European Union or the European Research Council. Neither the European Union nor the granting
authority can be held responsible for them. 

\bibliographystyle{elsarticle-num}
\bibliography{cas-refs}

@article{tabor1977surface,
  title={Surface forces and surface interactions},
  author={Tabor, D},
  journal={Journal of Colloid and Interface Science},
  volume={58},
  number={1},
  pages={2--13},
  year={1977},
  doi={10.1016/0021-9797(77)90366-6}
}

@article{kingma2014adam,
  title={Adam: A method for stochastic optimization},
  author={Kingma, Diederik P and Ba, Jimmy},
  journal={arXiv preprint arXiv:1412.6980},
  year={2014},
  doi={10.48550/arXiv.1412.6980}
}

@article{ziemer2025impact,
  title={Impact of single and combined space environment factors on the performance of elastomer micropatterned dry adhesives},
  author={Ziemer, Lennart and Garzke, MF and Hand, A and Ben-Larbi, MK and Stoll, E and Tighe, AP},
  journal={Polymer Testing},
  volume={149},
  pages={108866},
  year={2025},
  publisher={Elsevier}
}

@article{nguyen2026machine,
  title={Machine learning for adhesion assessment and prediction},
  author={Nguyen, Dang Nhat and Kaminski, Daniel and Subramanian, Priya and Bateman, Stuart and Le, Tu C},
  journal={Journal of Adhesion Science and Technology},
  pages={1--31},
  year={2026},
  publisher={Taylor \& Francis},
  doi={10.1080/01694243.2026.2687662}
}

@inproceedings{cuturi2017soft,
  title={Soft-dtw: a differentiable loss function for time-series},
  author={Cuturi, Marco and Blondel, Mathieu},
  booktitle={International conference on machine learning},
  pages={894--903},
  year={2017},
  organization={PMLR}
}

@article{giorgino2009computing,
  title={Computing and visualizing dynamic time warping alignments in R: the dtw package},
  author={Giorgino, Toni},
  journal={Journal of statistical Software},
  volume={31},
  pages={1--24},
  year={2009},
  doi={10.18637/jss.v031.i07}
}

@inproceedings{lin2003symbolic,
  title={A symbolic representation of time series, with implications for streaming algorithms},
  author={Lin, Jessica and Keogh, Eamonn and Lonardi, Stefano and Chiu, Bill},
  booktitle={Proceedings of the 8th ACM SIGMOD workshop on Research issues in data mining and knowledge discovery},
  pages={2--11},
  year={2003},
  doi={10.1145/882082.882086}
}

@article{stender2021deep,
  title={Deep learning for brake squeal: Brake noise detection, characterization and prediction},
  author={Stender, Merten and Tiedemann, Merten and Spieler, David and Schoepflin, Daniel and Hoffmann, Norbert and Oberst, Sebastian},
  journal={Mechanical Systems and Signal Processing},
  volume={149},
  pages={107181},
  year={2021},
  publisher={Elsevier},
  doi={10.1016/j.ymssp.2020.107181}
}

@article{humfeld2021machine,
  title={A machine learning framework for real-time inverse modeling and multi-objective process optimization of composites for active manufacturing control},
  author={Humfeld, Keith D and Gu, Dawei and Butler, Geoffrey A and Nelson, Karl and Zobeiry, Navid},
  journal={Composites Part B: Engineering},
  volume={223},
  pages={109150},
  year={2021},
  publisher={Elsevier},
  doi={10.1016/j.compositesb.2021.109150}
}

@article{sattari2020prediction,
  title={Prediction of nanoscale friction for two-dimensional materials using a machine learning approach},
  author={Sattari Baboukani, Behnoosh and Ye, Zhijiang and G Reyes, Kristofer and Nalam, Prathima C},
  journal={Tribology Letters},
  volume={68},
  pages={1--14},
  year={2020},
  publisher={Springer},
  doi={10.1007/s11249-020-01294-w}
}

@article{geier2023machine,
  title={Machine learning-based state maps for complex dynamical systems: applications to friction-excited brake system vibrations},
  author={Geier, Charlotte and Hamdi, Sa{\"\i}d and Chancelier, Thierry and Dufr{\'e}noy, Philippe and Hoffmann, Norbert and Stender, Merten},
  journal={Nonlinear dynamics},
  volume={111},
  number={24},
  pages={22137--22151},
  year={2023},
  publisher={Springer},
  doi={10.1007/s11071-023-08739-6}
}

@article{sahin2024solving,
  title={Solving forward and inverse problems of contact mechanics using physics-informed neural networks},
  author={Sahin, Tarik and von Danwitz, Max and Popp, Alexander},
  journal={Advanced Modeling and Simulation in Engineering Sciences},
  volume={11},
  number={1},
  pages={11},
  year={2024},
  publisher={Springer},
  doi={10.1186/s40323-024-00265-3}
}

@article{maugis1992adhesion,
  title={Adhesion of spheres: the JKR-DMT transition using a Dugdale model},
  author={Maugis, Daniel},
  journal={Journal of colloid and interface science},
  volume={150},
  number={1},
  pages={243--269},
  year={1992},
  publisher={Elsevier},
  doi={10.1016/0021-9797(92)90285-t}
}

@article{greenwood1997adhesion,
  title={Adhesion of elastic spheres},
  author={Greenwood, JA1455332},
  journal={Proceedings of the Royal Society of London. Series A: Mathematical, Physical and Engineering Sciences},
  volume={453},
  number={1961},
  pages={1277--1297},
  year={1997},
  publisher={The Royal Society},
  doi={10.1098/rspa.1997.0070}
}

@article{johnson1997adhesion,
  title={An adhesion map for the contact of elastic spheres},
  author={Johnson, KL and Greenwood, JA},
  journal={Journal of colloid and interface science},
  volume={192},
  number={2},
  pages={326--333},
  year={1997},
  publisher={Elsevier},
  doi={10.1006/jcis.1997.4984}
}

@article{shojaeifard2025machine,
  title={Machine learning-based optimal design of fibrillar adhesives},
  author={Shojaeifard, Mohammad and Ferraresso, Matteo and Lucantonio, Alessandro and Bacca, Mattia},
  journal={Journal of the Royal Society Interface},
  volume={22},
  number={223},
  pages={20240636},
  year={2025},
  publisher={The Royal Society},
  doi={10.1098/rsif.2024.0636}
}

@article{mackay2023informed,
  title={Informed machine learning methods for application in engineering: A review},
  author={Mackay, Calum Torin and Nowell, David},
  journal={Proceedings of the Institution of Mechanical Engineers, Part C: Journal of Mechanical Engineering Science},
  volume={237},
  number={24},
  pages={5801--5818},
  year={2023},
  publisher={SAGE Publications Sage UK: London, England},
  doi={10.1177/09544062231164575}
}

@article{giordano2024mechanochromic,
  title={Mechanochromic suction cups for local stress detection in soft robotics},
  author={Giordano, Goffredo and Scharff, Rob Bernardus Nicolaas and Carlotti, Marco and Gagliardi, Mariacristina and Filippeschi, Carlo and Mondini, Alessio and Papangelo, Antonio and Mazzolai, Barbara},
  journal={Advanced Intelligent Systems},
  pages={2400254},
  year={2024},
  publisher={Wiley Online Library},
  doi={10.1002/aisy.202400254}
}

@article{derjaguin1975effect,
  title={Effect of contact deformations on the adhesion of particles},
  author={Derjaguin, Boris V and Muller, Vladimir M and Toporov, Yu P},
  journal={Journal of Colloid and interface science},
  volume={53},
  number={2},
  pages={314--326},
  year={1975},
  publisher={Elsevier},
  doi={10.1016/0021-9797(75)90018-1}
}

@article{popov2021note,
  title={A Note by KL Johnson on the History of the JKR Theory},
  author={Popov, Valentin L},
  journal={Tribology Letters},
  volume={69},
  number={4},
  pages={132},
  year={2021},
  publisher={Springer},
  doi={10.1007/s11249-021-01511-0}
}

@article{johnson1971surface,
  title={Surface energy and the contact of elastic solids},
  author={Johnson, Kenneth Langstreth and Kendall, Kevin and Roberts, AAD},
  journal={Proceedings of the royal society of London. A. mathematical and physical sciences},
  volume={324},
  number={1558},
  pages={301--313},
  year={1971},
  publisher={The Royal Society London},
  doi={10.1098/rspa.1971.0141}
}

@article{liang2024autopeel,
  title={AutoPeel: Adhesion-aware Safe Peeling Trajectory Optimization for Robotic Wound Care},
  author={Liang, Xiao and Zhang, Youcheng and Liu, Fei and Richter, Florian and Yip, Michael},
  journal={arXiv preprint arXiv:2409.14282},
  year={2024},
  doi={10.48550/arXiv.2409.14282}
}

@article{ji2019apple,
  title={Apple viscoelastic complex model for bruise damage analysis in constant velocity grasping by gripper},
  author={Ji, Wei and Qian, Zhijie and Xu, Bo and Chen, Guangyu and Zhao, Dean},
  journal={Computers and Electronics in Agriculture},
  volume={162},
  pages={907--920},
  year={2019},
  publisher={Elsevier},
  doi={10.1016/j.compag.2019.05.022}
}

@article{guo2024interaction,
  title={Interaction between eye movements and adhesion of extraocular muscles},
  author={Guo, Hongmei and Lan, Yunfei and Gao, Zhipeng and Zhang, Chenxi and Zhang, Liping and Li, Xiaona and Lin, Jianying and Elsheikh, Ahmed and Chen, Weiyi},
  journal={Acta Biomaterialia},
  volume={176},
  pages={304--320},
  year={2024},
  publisher={Elsevier},
  doi={10.1016/j.actbio.2024.01.028}
}

@article{liu2022switchable,
  title={Switchable adhesion: on-demand bonding and debonding},
  author={Liu, Ziyang and Yan, Feng},
  journal={Advanced Science},
  volume={9},
  number={12},
  pages={2200264},
  year={2022},
  publisher={Wiley Online Library},
  doi={10.1002/advs.202200264}
}

@article{souza2010multiscale,
  title={Multiscale modeling of impact on heterogeneous viscoelastic solids containing evolving microcracks},
  author={Souza, Flavio V and Allen, David H},
  journal={International Journal for Numerical Methods in Engineering},
  volume={82},
  number={4},
  pages={464--504},
  year={2010},
  publisher={Wiley Online Library},
  doi={10.1002/nme.2773}
}

@article{ahmad2024family,
  title={A family of minimum residual displacement methods as nonlinear solution schemes for equilibrium path-following in structural mechanics},
  author={Ahmad-Abad, Mostafa Salehi and Maghami, Ali and Ghalishooyan, Morteza and Shooshtari, Ahmad},
  journal={Computers \& Structures},
  volume={300},
  pages={107407},
  year={2024},
  publisher={Elsevier},
  doi={10.1016/j.compstruc.2024.107407}
}

@article{tao2023climbing,
  title={Climbing robots for manufacturing},
  author={Tao, Bo and Gong, Zeyu and Ding, Han},
  journal={National Science Review},
  volume={10},
  number={5},
  pages={nwad042},
  year={2023},
  publisher={Oxford University Press},
  doi={10.1093/nsr/nwad042}
}

@article{kim2023designing,
  title={Designing directional adhesive pillars using deep learning-based optimization, 3D printing, and testing},
  author={Kim, Yongtae and Yeo, Jinwook and Park, Kundo and Destr{\'e}e, Aymeric and Qin, Zhao and Ryu, Seunghwa},
  journal={Mechanics of Materials},
  volume={185},
  pages={104778},
  year={2023},
  publisher={Elsevier},
  doi={10.1016/j.mechmat.2023.104778}
}

@article{ciavarella2019role,
  title={The role of adhesion in contact mechanics},
  author={Ciavarella, Michele and Joe, J and Papangelo, Antonio and Barber, JR},
  journal={Journal of the Royal Society Interface},
  volume={16},
  number={151},
  pages={20180738},
  year={2019},
  publisher={The Royal Society},
  doi={10.1098/rsif.2018.0738}
}

@article{dayan2024machine,
  title={Machine Learning-Based Shear Optimal Adhesive Microstructures with Experimental Validation},
  author={Dayan, Cem Balda and Son, Donghoon and Aghakhani, Amirreza and Wu, Yingdan and Demir, Sinan Ozgun and Sitti, Metin},
  journal={Small},
  volume={20},
  number={2},
  pages={2304437},
  year={2024},
  publisher={Wiley Online Library},
  doi={10.1002/smll.202304437}
}

@article{son2021machine,
  title={Machine Learning-Based and Experimentally Validated Optimal Adhesive Fibril Designs},
  author={Son, Donghoon and Liimatainen, Ville and Sitti, Metin},
  journal={Small},
  volume={17},
  number={39},
  pages={2102867},
  year={2021},
  publisher={Wiley Online Library},
  doi={10.1002/smll.202102867}
}

@article{luo2022machine,
  title={Machine learning-based optimization of the design of composite pillars for dry adhesives},
  author={Luo, Aoyi and Zhang, Hang and Turner, Kevin T},
  journal={Extreme Mechanics Letters},
  volume={54},
  pages={101695},
  year={2022},
  publisher={Elsevier},
  doi={10.1016/j.eml.2022.101695}
}

@article{kim2020designing,
  title={Designing an adhesive pillar shape with deep learning-based optimization},
  author={Kim, Yongtae and Yang, Charles and Kim, Youngsoo and Gu, Grace X and Ryu, Seunghwa},
  journal={ACS applied materials \& interfaces},
  volume={12},
  number={21},
  pages={24458--24465},
  year={2020},
  publisher={ACS Publications},
  doi={10.1021/acsami.0c04123}
}

@article{carbone2022theory,
  title={Theory of viscoelastic adhesion and friction},
  author={Carbone, Giuseppe and Mandriota, Cosimo and Menga, Nicola},
  journal={Extreme Mechanics Letters},
  volume={56},
  pages={101877},
  year={2022},
  publisher={Elsevier},
  doi={10.1016/j.eml.2022.101877}
}

@article{tricarico2025enhancement,
  title={Enhancement of adhesion strength through microvibrations: Modeling and experiments},
  author={Tricarico, Michele and Ciavarella, Michele and Papangelo, Antonio},
  journal={Journal of the Mechanics and Physics of Solids},
  volume={196},
  pages={106020},
  year={2025},
  publisher={Elsevier},
  doi={10.1016/j.jmps.2024.106020}
}

@article{ciavarella2025dynamic,
  title={On the dynamic JKR adhesion problem},
  author={Ciavarella, Michele and Tricarico, Michele and Papangelo, Antonio},
  journal={Mechanics of Materials},
  volume={202},
  pages={105252},
  year={2025},
  publisher={Elsevier},
  doi={10.1016/j.mechmat.2025.105252}
}

@article{guo2021artificial,
  title={Artificial intelligence and machine learning in design of mechanical materials},
  author={Guo, Kai and Yang, Zhenze and Yu, Chi-Hua and Buehler, Markus J},
  journal={Materials Horizons},
  volume={8},
  number={4},
  pages={1153--1172},
  year={2021},
  publisher={Royal society of chemistry},
  doi={10.1039/d0mh01451f}
}

@article{mandriota2024adhesive,
  title={Adhesive contact mechanics of viscoelastic materials},
  author={Mandriota, Cosimo and Menga, Nicola and Carbone, Giuseppe},
  journal={International Journal of Solids and Structures},
  volume={290},
  pages={112685},
  year={2024},
  publisher={Elsevier},
  doi={10.1016/j.ijsolstr.2024.112685}
}

@article{afferrante2022effective,
  title={On the effective surface energy in viscoelastic Hertzian contacts},
  author={Afferrante, Luciano and Violano, Guido},
  journal={Journal of the Mechanics and Physics of Solids},
  volume={158},
  pages={104669},
  year={2022},
  publisher={Elsevier},
  doi={10.1016/j.jmps.2021.104669}
}

@article{bonfanti2020fractional,
  title={Fractional viscoelastic models for power-law materials},
  author={Bonfanti, Alessandra and Kaplan, Jonathan Louis and Charras, Guillaume and Kabla, Alexandre},
  journal={Soft Matter},
  volume={16},
  number={26},
  pages={6002--6020},
  year={2020},
  publisher={Royal Society of Chemistry},
  doi={10.1039/d0sm00354a}
}

@book{christensen2012theory,
  title={Theory of viscoelasticity: an introduction},
  author={Christensen, Richard},
  year={2012},
  publisher={Elsevier}
}

@article{ciavarella2021comparison,
  title={A comparison of crack propagation theories in viscoelastic materials},
  author={Ciavarella, Michele and Cricr{\`\i}, Gabriele and McMeeking, R},
  journal={Theoretical and applied fracture mechanics},
  volume={116},
  pages={103113},
  year={2021},
  publisher={Elsevier},
  doi={10.1016/j.tafmec.2021.103113}
}

@article{feng2000contact,
  title={Contact behavior of spherical elastic particles: a computational study of particle adhesion and deformations},
  author={Feng, James Q},
  journal={Colloids and Surfaces A: Physicochemical and Engineering Aspects},
  volume={172},
  number={1-3},
  pages={175--198},
  year={2000},
  publisher={Elsevier},
  doi={10.1016/s0927-7757(00)00580-x}
}

@article{greenwood2004theory,
  title={The theory of viscoelastic crack propagation and healing},
  author={Greenwood, JA},
  journal={Journal of Physics D: Applied Physics},
  volume={37},
  number={18},
  pages={2557},
  year={2004},
  publisher={IOP Publishing},
  doi={10.1088/0022-3727/37/18/011}
}

@article{greenwood1981mechanics,
  title={The mechanics of adhesion of viscoelastic solids},
  author={Greenwood, JA and Johnson, KL},
  journal={Philosophical Magazine A},
  volume={43},
  number={3},
  pages={697--711},
  year={1981},
  publisher={Taylor \& Francis},
  doi={10.1080/01418618108240402}
}

@book{johnson1987contact,
  author    = {Johnson, Kenneth L.},
  title     = {Contact Mechanics},
  year      = {1987},
  publisher = {Cambridge University Press},
  address   = {Cambridge},
  doi={10.1017/cbo9781139171731}
}

@article{maghami2024viscoelastic,
  title={Viscoelastic amplification of the pull-off stress in the detachment of a rigid flat punch from an adhesive soft viscoelastic layer},
  author={Maghami, Ali and Tricarico, Michele and Ciavarella, Michele and Papangelo, Antonio},
  journal={Engineering Fracture Mechanics},
  volume={298},
  pages={109898},
  year={2024},
  publisher={Elsevier},
  doi={10.1016/j.engfracmech.2024.109898}
}

@article{eshkofti2024modified,
  title={The modified physics-informed neural network (PINN) method for the thermoelastic wave propagation analysis based on the Moore-Gibson-Thompson theory in porous materials},
  author={Eshkofti, Katayoun and Hosseini, Seyed Mahmoud},
  journal={Composite Structures},
  volume={348},
  pages={118485},
  year={2024},
  publisher={Elsevier},
  doi={10.1016/j.compstruct.2024.118485}
}

@article{kellner2019establishing,
  title={Establishing a common database of ice experiments and using machine learning to understand and predict ice behavior},
  author={Kellner, Leon and Stender, Merten and Herrnring, Hauke and Ehlers, S{\"o}ren and Hoffmann, Norbert and H{\o}yland, Knut V and others},
  journal={Cold regions science and technology},
  volume={162},
  pages={56--73},
  year={2019},
  publisher={Elsevier},
  doi={10.1016/j.coldregions.2019.02.007}
}

@article{javadi2022deep,
  title={A deep learning approach based on a data-driven tool for classification and prediction of thermoelastic wave's band structures for phononic crystals},
  author={Javadi, Shirin and Maghami, Ali and Hosseini, Seyed Mahmoud},
  journal={Mechanics of Advanced Materials and Structures},
  volume={29},
  number={27},
  pages={6612--6625},
  year={2022},
  publisher={Taylor \& Francis},
  doi={10.1080/15376494.2021.1983088}
}

@article{kalliorinne2021artificial,
  title={Artificial neural network architecture for prediction of contact mechanical response},
  author={Kalliorinne, Kalle and Larsson, Roland and P{\'e}rez-R{\`a}fols, Francesc and Liwicki, Marcus and Almqvist, Andreas},
  journal={Frontiers in Mechanical Engineering},
  volume={6},
  pages={579825},
  year={2021},
  publisher={Frontiers Media SA},
  doi={10.3389/fmech.2020.579825}
}

@article{perera2023generalized,
  title={A generalized machine learning framework for brittle crack problems using transfer learning and graph neural networks},
  author={Perera, Roberto and Agrawal, Vinamra},
  journal={Mechanics of Materials},
  volume={181},
  pages={104639},
  year={2023},
  publisher={Elsevier},
  doi={10.1016/j.mechmat.2023.104639}
}

@article{li2022machine,
  title={Machine learning-based prediction of fracture toughness and path in the presence of micro-defects},
  author={Li, Xiaotao and Zhang, Xu and Feng, Wei and Wang, Qingyuan},
  journal={Engineering Fracture Mechanics},
  volume={276},
  pages={108900},
  year={2022},
  publisher={Elsevier},
  doi={10.1016/j.engfracmech.2022.108900}
}

@article{yi2024mechanics,
  title={Mechanics-informed, model-free symbolic regression framework for solving fracture problems},
  author={Yi, Ruibang and Georgiou, Dimitrios and Liu, Xing and Athanasiou, Christos E},
  journal={Journal of the Mechanics and Physics of Solids},
  pages={105916},
  year={2024},
  publisher={Elsevier},
  doi={10.1016/j.jmps.2024.105916}
}

@article{athanasiou2023integrated,
  title={Integrated simulation, machine learning, and experimental approach to characterizing fracture instability in indentation pillar-splitting of materials},
  author={Athanasiou, Christos E and Liu, Xing and Zhang, Boyu and Cai, Truong and Ramirez, Cristina and Padture, Nitin P and Lou, Jun and Sheldon, Brian W and Gao, Huajian},
  journal={Journal of the Mechanics and Physics of Solids},
  volume={170},
  pages={105092},
  year={2023},
  publisher={Elsevier},
  doi={10.1016/j.jmps.2022.105092}
}

@article{wang2021machine,
  title={Machine learning approaches to rock fracture mechanics problems: Mode-I fracture toughness determination},
  author={Wang, Yun-Teng and Zhang, Xiang and Liu, Xian-Shan},
  journal={Engineering Fracture Mechanics},
  volume={253},
  pages={107890},
  year={2021},
  publisher={Elsevier},
  doi={10.1016/j.engfracmech.2021.107890}
}

@article{motiwale2024neural,
  title={A neural network finite element approach for high speed cardiac mechanics simulations},
  author={Motiwale, Shruti and Zhang, Wenbo and Feldmeier, Reese and Sacks, Michael S},
  journal={Computer Methods in Applied Mechanics and Engineering},
  volume={427},
  pages={117060},
  year={2024},
  publisher={Elsevier},
  doi={10.1016/j.cma.2024.117060}
}

@article{goodbrake2024neural,
  title={A neural network finite element method for contact mechanics},
  author={Goodbrake, Christian and Motiwale, Shruti and Sacks, Michael S},
  journal={Computer Methods in Applied Mechanics and Engineering},
  volume={419},
  pages={116671},
  year={2024},
  publisher={Elsevier},
  doi={10.1016/j.cma.2023.116671}
}

@article{papangelo2020numerical,
  title={A numerical study on roughness-induced adhesion enhancement in a sphere with an axisymmetric sinusoidal waviness using Lennard--Jones interaction law},
  author={Papangelo, Antonio and Ciavarella, Michele},
  journal={Lubricants},
  volume={8},
  number={9},
  pages={90},
  year={2020},
  publisher={MDPI},
  doi={10.3390/lubricants8090090}
}

@article{papangelo2023detachment,
  title={Detachment of a rigid flat punch from a viscoelastic material},
  author={Papangelo, Antonio and Ciavarella, Michele},
  journal={Tribology letters},
  volume={71},
  number={2},
  pages={48},
  year={2023},
  publisher={Springer},
  doi={10.1007/s11249-023-01720-9}
}

@article{persson2005crack,
  title={Crack propagation in viscoelastic solids},
  author={Persson, BNJ and Brener, EA},
  journal={Physical Review E},
  volume={71},
  number={3},
  pages={036123},
  year={2005},
  publisher={APS},
  doi={10.1103/physreve.71.036123}
}

@article{maghami2024bulk,
  title={Bulk and fracture process zone contribution to the rate-dependent adhesion amplification in viscoelastic broad-band materials},
  author={Maghami, Ali and Wang, Qingao and Tricarico, Michele and Ciavarella, Michele and Li, Qunyang and Papangelo, Antonio},
  journal={Journal of the Mechanics and Physics of Solids},
  volume={193},
  pages={105844},
  year={2024},
  publisher={Elsevier},
  doi={10.1016/j.jmps.2024.105844}
}

@misc{papangelo2024jmpsdata,
  author={Papangelo, Antonio},
  title={Data for ``Bulk and fracture process zone contribution to the rate-dependent adhesion amplification in viscoelastic broad-band materials''},
  year={2024},
  publisher={Zenodo},
  doi={10.5281/zenodo.13358696},
  url={https://doi.org/10.5281/zenodo.13358696}
}

@article{schapery1975theory1,
  title={A theory of crack initiation and growth in viscoelastic media: I. Theoretical development},
  author={Schapery, Richard A},
  journal={International Journal of fracture},
  volume={11},
  pages={141--159},
  year={1975},
  publisher={Springer},
  doi={10.1007/BF00034721}
}

@article{violano2021jkr,
  title={A JKR-like solution for viscoelastic adhesive contacts},
  author={Violano, Guido and Chateauminois, Antoine and Afferrante, Luciano},
  journal={Frontiers in Mechanical Engineering},
  volume={7},
  pages={664486},
  year={2021},
  publisher={Frontiers Media SA},
  doi={10.3389/fmech.2021.664486}
}

@article{george2022closing,
  title={Closing the control loop with time-variant embedded soft sensors and recurrent neural networks},
  author={George Thuruthel, Thomas and Gardner, Paul and Iida, Fumiya},
  journal={Soft Robotics},
  volume={9},
  number={6},
  pages={1167--1176},
  year={2022},
  publisher={Mary Ann Liebert, Inc., publishers 140 Huguenot Street, 3rd Floor New~…},
  doi={10.1089/soro.2021.0012}
}

@article{hinrichsen2024using,
  title={Using dropout based active learning and surrogate models in the inverse viscoelastic parameter identification of human brain tissue},
  author={Hinrichsen, Jan and Ferlay, Carl and Reiter, Nina and Budday, Silvia},
  journal={Frontiers in Physiology},
  volume={15},
  pages={1321298},
  year={2024},
  publisher={Frontiers Media SA},
  doi={10.3389/fphys.2024.1321298}
}

@article{karami2023real,
  title={Real-time simulation of viscoelastic tissue behavior with physics-guided deep learning},
  author={Karami, Mohammad and Lombaert, Herv{\'e} and Rivest-H{\'e}nault, David},
  journal={Computerized Medical Imaging and Graphics},
  volume={104},
  pages={102165},
  year={2023},
  publisher={Elsevier},
  doi={10.1016/j.compmedimag.2022.102165}
}

@article{koeppe2022explainable,
  title={Explainable artificial intelligence for mechanics: physics-explaining neural networks for constitutive models},
  author={Koeppe, Arnd and Bamer, Franz and Selzer, Michael and Nestler, Britta and Markert, Bernd},
  journal={Frontiers in Materials},
  volume={8},
  pages={824958},
  year={2022},
  publisher={Frontiers Media SA},
  doi={10.3389/fmats.2021.824958}
}

@article{chen2021recurrent,
  title={Recurrent neural networks (RNNs) learn the constitutive law of viscoelasticity},
  author={Chen, Guang},
  journal={Computational Mechanics},
  volume={67},
  number={3},
  pages={1009--1019},
  year={2021},
  publisher={Springer},
  doi={10.1007/s00466-021-01981-y}
}

@article{li2023modeling,
  title={Modeling of magnetorheological dampers based on a dual-flow neural network with efficient channel attention},
  author={Li, Jiahao and Luo, Jiayang and Zhang, Feng and Zhou, Wei and Wei, Xin and Liao, Changrong and Shou, Mengjie},
  journal={Smart Materials and Structures},
  volume={32},
  number={10},
  pages={105006},
  year={2023},
  publisher={IOP Publishing},
  doi={10.1088/1361-665x/acf016}
}

@article{wang2024drift,
  title={Drift-Aware Feature Learning Based on Autoencoder Preprocessing for Soft Sensors},
  author={Wang, Junming and Shu, Jing and Alam, Md Masruck and Gao, Zhaoli and Li, Zheng and Tong, Raymond Kai-Yu},
  journal={Advanced Intelligent Systems},
  volume={6},
  number={3},
  pages={2300486},
  year={2024},
  publisher={Wiley Online Library},
  doi={10.1002/aisy.202300486}
}

@article{yan2021machine,
  title={Machine learning assisted discovery of new thermoset shape memory polymers based on a small training dataset},
  author={Yan, Cheng and Feng, Xiaming and Wick, Collin and Peters, Andrew and Li, Guoqiang},
  journal={Polymer},
  volume={214},
  pages={123351},
  year={2021},
  publisher={Elsevier},
  doi={10.1016/j.polymer.2020.123351}
}

@article{lu2026llm,
  title={A LLM-inspired experimental-data-driven framework for viscoelastic soft structures},
  author={Lu, Yicheng and Hu, Xinjie and Pu, Yichen and Yu, Zefeng and An, Ning and Tang, Shan and Guo, Xu},
  journal={International Journal of Mechanical Sciences},
  pages={111696},
  year={2026},
  publisher={Elsevier},
  doi={10.1016/j.ijmecsci.2026.111696}
}

@article{maghami2025pull,
  title={Pull-off force prediction in viscoelastic adhesive Hertzian contact by physics augmented machine learning},
  author={Maghami, Ali and Stender, Merten and Papangelo, Antonio},
  journal={International Journal of Solids and Structures},
  volume={322},
  pages={113584},
  year={2025},
  publisher={Elsevier},
  doi={10.1016/j.ijsolstr.2025.113584}
}

@article{hochreiter1997long,
  title={Long short-term memory},
  author={Hochreiter, Sepp and Schmidhuber, J{\"u}rgen},
  journal={Neural computation},
  volume={9},
  number={8},
  pages={1735--1780},
  year={1997},
  publisher={MIT press},
  doi={10.1162/neco.1997.9.8.1735}
}

@article{bai2018empirical,
  title={An empirical evaluation of generic convolutional and recurrent networks for sequence modeling},
  author={Bai, Shaojie and Kolter, J Zico and Koltun, Vladlen},
  journal={arXiv preprint arXiv:1803.01271},
  year={2018},
  doi={10.48550/arXiv.1803.01271}
}

@article{williams1964structural,
  title={Structural analysis of viscoelastic materials},
  author={Williams, M Lr},
  journal={AIAA journal},
  volume={2},
  number={5},
  pages={785--808},
  year={1964},
  doi={10.2514/3.2447}
}

@misc{PAPANGELO2026,
title = {Adhesive Single and Multi-asperity Contacts},
author = {A. Papangelo and M. Tricarico and A. Maghami},
howpublished = {Reference Module in Materials Science and Materials Engineering},
publisher = {Elsevier},
year = {2026},
isbn = {978-0-12-803581-8},
doi = {10.1016/B978-0-443-30138-4.00009-1},
}

@article{revAbushattal2026,
  title = {Nonlinear Receding Contact Mechanics of Functionally Graded Layers for Aerospace Structures: A Symmetry-Based Analytical and FEM Study},
  author = {Abushattal, Ahmad and Terzi, Merve and Eyübo\u{g}lu, Ayşegül and Yaylac{\i}, Murat and Sekban, Dursun Murat and Nay{\i}r, Safa and Yaylac{\i}, Ecren Uzun and Dewi, Deshinta Arrova and Birinci, Ahmet},
  journal = {Symmetry},
  volume = {18},
  number = {2},
  pages = {378},
  year = {2026},
  doi = {10.3390/sym18020378}
}

@article{revZhang2026,
  title = {Axisymmetric Adhesive Contact Mechanics of Piezoelectric Semiconductors},
  author = {Zhang, Yali and Zhou, Yueting and Ding, Shenghu},
  journal = {Mechanics of Materials},
  volume = {218},
  pages = {105675},
  year = {2026},
  doi = {10.1016/j.mechmat.2026.105675}
}

@article{revZhang2025,
  title = {Periodic Partial Slip Contact of Thermoelectric Materials under Flat Punches or Cylindrical Punches},
  author = {Zhang, Yali and Zhou, Yueting and Yang, Wenxian and Wang, Wenshuai and Ding, Shenghu},
  journal = {Acta Mechanica Sinica},
  volume = {41},
  pages = {424676},
  year = {2025},
  doi = {10.1007/s10409-024-24676-x}
}

@article{revDuan2025,
  title = {Sliding Adhesion Contact Problem of One-Dimensional Hexagonal Piezoelectric Quasicrystals Half-Space},
  author = {Duan, Yapeng and Du, Jiale and Zhang, Xin and Ding, Shenghu and Huang, Rukai},
  journal = {Acta Mechanica},
  volume = {236},
  number = {10},
  pages = {6023--6044},
  year = {2025},
  doi = {10.1007/s00707-025-04454-y}
}

@article{revMa2025,
  title = {Three Dimensional Thermally Frictional Adhesive Contact Problem of Quasicrystals Materials},
  author = {Ma, Lili and Li, Xing and Li, Yanxin and Liu, Junqiao and Pan, Bo and Tian, Yun and Ding, Shenghu and Zhang, Xin},
  journal = {Applied Mathematical Modelling},
  volume = {138},
  pages = {115819},
  year = {2025},
  doi = {10.1016/j.apm.2024.115819}
}

@article{revYaylaci2025,
  title = {Evaluation of the Contact Problem of Two Layers One of Functionally Graded, Loaded by Circular Rigid Block and Resting on a Pasternak Foundation by Analytical and Numerical (FEM and MLP) Methods},
  author = {Yaylac{\i}, Murat and Yaz{\i}c{\i}o\u{g}lu, Aleyna and Yaylac{\i}, Ecren Uzun and Terzi, Merve and Birinci, Ahmet},
  journal = {Archive of Applied Mechanics},
  volume = {95},
  number = {4},
  pages = {78},
  year = {2025},
  doi = {10.1007/s00419-025-02787-7}
}

@article{revOner2025,
  title = {Tri-Method Analysis of Contact Mechanics in Orthotropic-Isotropic Materials},
  author = {Öner, Erdal and Oktay, Mine Gül and Yaylac{\i}, Ecren Uzun and Yaylac{\i}, Murat and Mirzalo\u{g}lu, \.{I}rem},
  journal = {Archive of Applied Mechanics},
  volume = {95},
  pages = {116},
  year = {2025},
  doi = {10.1007/s00419-025-02832-5}
}

@article{tricarico2026influence,
  title={Influence of material and geometrical parameters on the adhesive performance of vibration-modulated soft contacts},
  author={Tricarico, Michele and Shiferaw, AY and Papangelo, Antonio},
  journal={European Journal of Mechanics-A/Solids},
  pages={106130},
  year={2026},
  publisher={Elsevier},
  doi={10.1016/j.euromechsol.2026.106130}
}

@article{mainardi2011creep,
  title={Creep, relaxation and viscosity properties for basic fractional models in rheology},
  author={Mainardi, Francesco and Spada, Giorgio},
  journal={The European Physical Journal Special Topics},
  volume={193},
  number={1},
  pages={133--160},
  year={2011},
  publisher={Springer},
  doi={10.1140/epjst/e2011-01387-1}
}

@incollection{sahin2024physics,
  title={Physics-informed neural networks for solving contact problems in three dimensions},
  author={Sahin, Tarik and Wolff, Daniel and Popp, Alexander},
  booktitle={Advances and Challenges in Computational Mechanics},
  pages={419--431},
  year={2026},
  publisher={Springer},
  doi={10.1007/978-3-031-93213-7\_33}
}

@article{shui2020rapid,
  title={Rapid and continuous regulating adhesion strength by mechanical micro-vibration},
  author={Shui, Langquan and Jia, Laibing and Li, Hangbo and Guo, Jiaojiao and Guo, Ziyu and Liu, Yilun and Liu, Ze and Chen, Xi},
  journal={Nature communications},
  volume={11},
  number={1},
  pages={1583},
  year={2020},
  publisher={Nature Publishing Group UK London},
  doi={10.1038/s41467-020-15447-x}
}

@article{dumoulin2018featurewise,
  title={Feature-wise transformations},
  author={Dumoulin, Vincent and Perez, Ethan and Schucher, Nathan and Strub, Florian and de Vries, Harm and Courville, Aaron and Bengio, Yoshua},
  journal={Distill},
  volume={3},
  number={7},
  pages={e11},
  year={2018},
  doi={10.23915/distill.00011}
}

@inproceedings{perez2018film,
  title={{FiLM}: Visual Reasoning with a General Conditioning Layer},
  author={Perez, Ethan and Strub, Florian and de Vries, Harm and Dumoulin, Vincent and Courville, Aaron},
  booktitle={Proceedings of the AAAI Conference on Artificial Intelligence},
  volume={32},
  pages={3942--3951},
  year={2018},
  doi={10.1609/aaai.v32i1.11671}
}

@article{arevalo2017gated,
  title={Gated Multimodal Units for Information Fusion},
  author={Arevalo, John and Solorio, Thamar and Montes-y-G{\'o}mez, Manuel and Gonz{\'a}lez, Fabio A.},
  journal={arXiv preprint arXiv:1702.01992},
  year={2017},
  eprint={1702.01992},
  archivePrefix={arXiv},
  primaryClass={stat.ML},
  doi={10.48550/arXiv.1702.01992}
}
\appendix
\section{Protocol definitions and BEM/Persson--Brener reference mapping}\label{sec:appendix_fig2}
This appendix documents the protocols for the six representative trajectories in Figure~\ref{fig:force_vs_time_unsteady} and the complete details and normalization of the relaxed short-range verification in Figure~\ref{fig:tier1d_pareto}(c). Table~\ref{tab:fig2_protocol_parameters} lists the loading rate, dwell duration, unloading rate, Tabor parameter, and peak indentation used in Figure~\ref{fig:force_vs_time_unsteady}(a)--(f), with the common initial indentation $\widehat{\delta}_0=-\mu\pi^{2/3}$.

\begin{table}[h]
    \centering
    \caption{{Loading-protocol parameters for the representative trajectories shown in Figure~\ref{fig:force_vs_time_unsteady}(a)--(f). The dimensionless loading and unloading velocities $\widehat{v}_L$ and $\widehat{v}_U$ are defined in Eq.~\protect\ref{eq:dlesspar}; $\widehat{t}_D$ is the dwell-phase duration, and $\widehat{\delta}_l$ is the peak normalized indentation at the onset of unloading (see Figure~\ref{fig:load_and_unloading}(e)).}}
    \label{tab:fig2_protocol_parameters}
    \small
    \setlength{\tabcolsep}{4pt}
    \begin{tabularx}{\linewidth}{@{}L{3.1cm}CCCCC@{}}
        \toprule
        \shortstack[l]{Figure~\ref{fig:force_vs_time_unsteady}\\panel}
            & \shortstack{Loading\\rate $\widehat{v}_L$}
            & \shortstack{Dwell time\\$\widehat{t}_D$}
            & \shortstack{Unloading\\rate $\widehat{v}_U$}
            & \shortstack{Tabor\\parameter $\mu$}
            & \shortstack{Peak indentation\\$\widehat{\delta}_l$} \\
        \midrule
        (a) & $1.00\times10^{3}$ & $8.88\times10^{-3}$ & $1.97\times10^{2}$ & $3.20$ & $2.85$ \\
        (b) & $1.00\times10^{3}$ & $2.07\times10^{-3}$ & $3.87\times10^{1}$ & $0.20$  & $34.4$ \\
        (c) & $3.38\times10^{2}$ & $1.00\times10^{-3}$ & $4.44$              & $2.49$  & $4.07$ \\
        (d) & $4.44$              & $1.00\times10^{-3}$ & $1.15\times10^{2}$ & $0.20$  & $2.85$ \\
        (e) & $0.508$             & $3.81\times10^{-2}$ & $0.873$             & $0.20$  & $11.8$ \\
        (f) & $0.508$             & $0.338$             & $2.58$              & $0.20$  & $70.1$ \\
        \bottomrule
    \end{tabularx}
\end{table}

The reference data in Figure~\ref{fig:tier1d_pareto}(c) are traced to two figures of Maghami et al.\ \cite{maghami2024bulk} and to their deposited dataset \cite{papangelo2024jmpsdata}. The red-square reference are the 14 out of 19-point blue-circle SLS series with modulus ratio $k=0.1$ in Figure~6 of Ref.~\cite{maghami2024bulk}, where normalized effective surface energy $\widehat{\Gamma}_{\mathrm{eff}}$ is plotted against the normalized crack velocity $\widetilde{v}_c$ defined in Section~\ref{sec:relaxed_short_range_verification}. The black solid curve is the PB solution for the same SLS material. The rate--velocity correspondence needed for the M1-concat points is taken from the blue-circle SLS series in the inset of Figure~7 of Ref.~\cite{maghami2024bulk}; that inset plots $\widetilde{v}_c$ against the PB-normalized unloading rate $\widetilde{v}_{\mathrm{U}}=v_U\tau_r/l_0$. The same numerical cases and marker identities are used in the main panel of Ref.~\cite{maghami2024bulk} Figure~7, and the deposited effective-surface-energy values for this SLS series are identical to those of the Ref.~\cite{maghami2024bulk} Figure~6 SLS series, providing a direct check of the identification.

According to  ~\cite{maghami2024bulk,persson2005crack} $l_0=E_0^*\Delta\gamma_0/[\pi(\alpha\sigma_0)^2]$ is the PB stress-based characteristic length, with $E_0^*$ the rubbery plane-strain modulus, $\Delta\gamma_0$ the thermodynamic surface energy, $\sigma_0$ the maximum tensile stress in the LJ interaction law and $\alpha$ an empirical parameter to relate the $\sigma_0$ in the LJ potential to the critical tensile strength $\sigma_c$ introduced by PB theory \cite{maghami2024bulk,persson2005crack}. Substitution of this PB rate into the definition of $\widehat{v}_U$ in Eq.~\ref{eq:dlesspar} for the SLS, gives
\begin{equation}
\widehat{v}_U=\frac{l_0}{h_0}\widetilde{v}_{\mathrm{U}}\approx4.252\,\widetilde{v}_{\mathrm{U}}.
\label{eq:jmps_rate_conversion}
\end{equation}
where from the definition of $l_0$ and from the deposited data of Figure~7 in  Ref.~\cite{maghami2024bulk}, one gets $l_0/h_0=\frac{\beta^2(R/h_0)^{1/2}}{\mu^{3/2}\pi\alpha^2}\approx4.252$, using the parameters $\beta=9\sqrt{3}/16$, $\alpha=\pi/9$, $\mu=3.24$, and $R/h_0=100$.

For each selected $\widetilde{v}_{\mathrm{U}}$ in Figure~7 of Ref.~\cite{maghami2024bulk} inset rate, Eq.~\ref{eq:jmps_rate_conversion} gives the unloading-velocity input $\widehat{v}_U$ supplied to M1-concat, while the corresponding normalized crack velocity $\widetilde{v}_c$ is read from the same deposited inset pair. Table~\ref{tab:fig5c_protocol_parameters} lists the fourteen selected pairs. Across these protocols, only $\widehat{v}_U$ varies; $\mu=3.24$, $\widehat{\delta}_l=80$, $\widehat{v}_L=1.3895$, and $\widehat{t}_D=3.0$ are held fixed to approach a highly preloaded, relaxed short-range state before unloading. Since the dataset was generated up to $\mu=3.2$, the M1-concat evaluation at $\mu=3.24$ is a controlled 1.25\% extrapolation.

\begin{table}[h]
    \centering
    \caption{Protocols used for the relaxed short-range verification in Figure~\protect\ref{fig:tier1d_pareto}(c). The unloading velocity $\widehat{v}_U$ is the only varied surrogate input. Each value is obtained from a deposited SLS retraction rate in the inset of Ref.~\cite{maghami2024bulk} Figure~7 through Eq.~\protect\ref{eq:jmps_rate_conversion}, and the corresponding normalized crack velocity $\widetilde{v}_c$ is taken from the same inset pair. Displayed values are rounded, whereas the calculations use the full deposited precision.}

    \label{tab:fig5c_protocol_parameters}
    \scriptsize
    \setlength{\tabcolsep}{2pt}
    \renewcommand{\arraystretch}{0.88}
    \begin{tabularx}{\linewidth}{@{}CCCCCCC@{}}
        \toprule
        \shortstack{Point}
            & \shortstack{Tabor parameter\\$\mu$}
            & \shortstack{Peak indentation\\$\widehat{\delta}_l$}
            & \shortstack{Loading rate\\$\widehat{v}_L$}
            & \shortstack{Dwell time\\$\widehat{t}_D$}
            & \shortstack{Unloading rate\\$\widehat{v}_U$}
            & \shortstack{Crack velocity\\from Ref.~\cite{maghami2024bulk}\\$\widetilde{v}_c$} \\
        \midrule
        1  & $3.24$ & $80$ & $1.3895$ & $3.0$ & $1.00\times10^{-1}$ & $6.00\times10^{-2}$ \\
        2  & $3.24$ & $80$ & $1.3895$ & $3.0$ & $1.78\times10^{-1}$ & $1.01\times10^{-1}$ \\
        3  & $3.24$ & $80$ & $1.3895$ & $3.0$ & $3.16\times10^{-1}$ & $1.69\times10^{-1}$ \\
        4  & $3.24$ & $80$ & $1.3895$ & $3.0$ & $5.62\times10^{-1}$ & $2.69\times10^{-1}$ \\
        5  & $3.24$ & $80$ & $1.3895$ & $3.0$ & $1.00$ & $4.21\times10^{-1}$ \\
        6  & $3.24$ & $80$ & $1.3895$ & $3.0$ & $5.62$ & $2.31$ \\
        7  & $3.24$ & $80$ & $1.3895$ & $3.0$ & $1.00\times10^{1}$ & $3.86$ \\
        8  & $3.24$ & $80$ & $1.3895$ & $3.0$ & $1.78\times10^{1}$ & $6.58$ \\
        9  & $3.24$ & $80$ & $1.3895$ & $3.0$ & $3.16\times10^{1}$ & $1.12\times10^{1}$ \\
        10 & $3.24$ & $80$ & $1.3895$ & $3.0$ & $5.62\times10^{1}$ & $1.98\times10^{1}$ \\
        11 & $3.24$ & $80$ & $1.3895$ & $3.0$ & $1.00\times10^{2}$ & $9.74\times10^{1}$ \\
        12 & $3.24$ & $80$ & $1.3895$ & $3.0$ & $1.78\times10^{2}$ & $3.23\times10^{2}$ \\
        13 & $3.24$ & $80$ & $1.3895$ & $3.0$ & $3.16\times10^{2}$ & $7.58\times10^{2}$ \\
        14 & $3.24$ & $80$ & $1.3895$ & $3.0$ & $5.62\times10^{2}$ & $1.51\times10^{3}$ \\
        \bottomrule
    \end{tabularx}
\end{table}

\subsection{Exact Persson \& Brener relation for the SLS}\label{app:pb_sls_relation}

For the SLS creep compliance in Eq.~\ref{eq:Ctgen}, let $C_0=1/E_0$ and $C_\infty=1/E_\infty=kC_0$. With the retardation-spectrum convention $C(t)=C_\infty+\int_0^\infty \tau^{-1}L(\tau)[1-\exp(-t/\tau)]\,d\tau$. Consequently, the PB spectrum integral for the normalized effective surface energy $\widehat{\Gamma}_{\mathrm{eff}}=\Delta\gamma_{\mathrm{eff}}/\Delta\gamma_0$ reduces to the exact implicit SLS relation \cite{persson2005crack,ciavarella2021comparison,maghami2024bulk}
\begin{equation}
\widehat{\Gamma}_{\mathrm{eff}}
=\left[
1-(1-k)\left(
\sqrt{1+\left(\frac{\widehat{\Gamma}_{\mathrm{eff}}}{2\pi\widetilde{v}_c}\right)^2}
-\frac{\widehat{\Gamma}_{\mathrm{eff}}}{2\pi\widetilde{v}_c}
\right)
\right]^{-1},
\label{eq:pb_sls_implicit}
\end{equation}
where $\widetilde{v}_c=v_c\tau_r/l_0$ is defined in Section~\ref{sec:relaxed_short_range_verification}. Equation~\ref{eq:pb_sls_implicit} is equivalent to the standard-material PB relation of Ref.~\cite{ciavarella2021comparison} after identifying its modulus-contrast parameter as $1-k$ and its normalized velocity as $2\pi\widehat{V}=\widetilde{v}_c$. An equivalent polynomial condition for the physical branch is
\begin{equation}
\begin{aligned}
&(1-k)\widehat{\Gamma}_{\mathrm{eff}}^3
+\left[\pi\widetilde{v}_c k(2-k)-(1-k)\right]
 \widehat{\Gamma}_{\mathrm{eff}}^2 \\
&\qquad -2\pi\widetilde{v}_c\widehat{\Gamma}_{\mathrm{eff}}
+\pi\widetilde{v}_c=0,
\end{aligned}
\label{eq:pb_sls_cubic}
\end{equation}
The physical root satisfies $1\leq\widehat{\Gamma}_{\mathrm{eff}}\leq1/k$, with $\widehat{\Gamma}_{\mathrm{eff}}=1$ at $\widetilde{v}_c=0$ and $\widehat{\Gamma}_{\mathrm{eff}}\rightarrow1/k$ as $\widetilde{v}_c\rightarrow\infty$. The same branch can be inverted exactly as
\begin{equation}
\widetilde{v}_c
=\frac{(1-k)\widehat{\Gamma}_{\mathrm{eff}}^2
(\widehat{\Gamma}_{\mathrm{eff}}-1)}
{\pi\left[(1-k)^2\widehat{\Gamma}_{\mathrm{eff}}^2
-(\widehat{\Gamma}_{\mathrm{eff}}-1)^2\right]}.
\label{eq:pb_sls_inverse}
\end{equation}

\section{Supplementary Architecture Figures}\label{sec:appendix_arch}

\begin{figure}
    \centering
\includegraphics[width=1\linewidth]{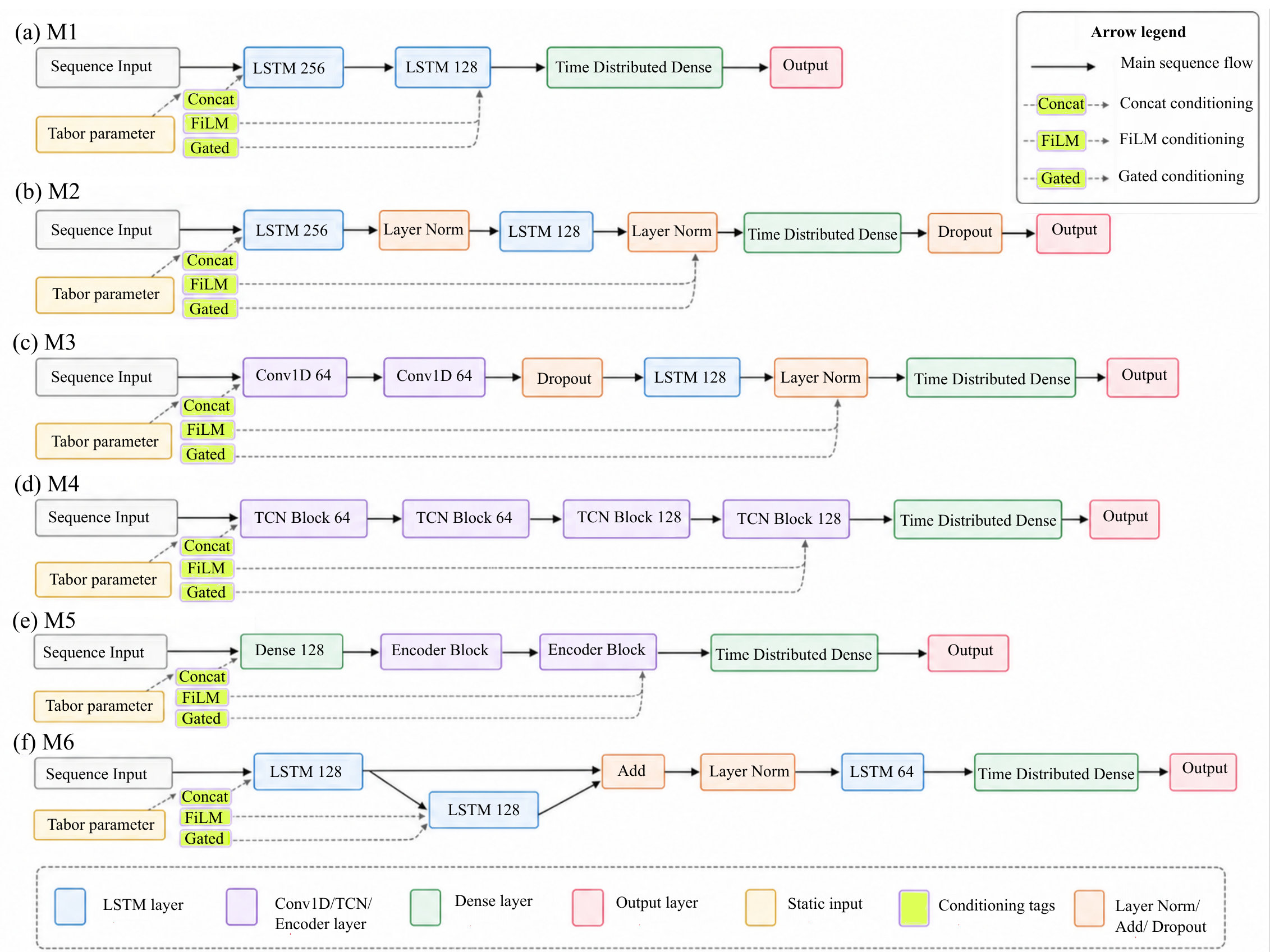}
    \caption{Neural network architectures for sequence modeling with Tabor parameter conditioning, ordered according to the paper labels and saved model indices. Solid arrows denote the main sequence/data flow, while dotted arrows denote Tabor-conditioning paths; the inset legend identifies the three conditioning mechanisms as [Concat], [FiLM], and [Gated]. (a) M1: LSTM base with LSTM(256) $\to$ LSTM(128) $\to$ TimeDistributed Dense. (b) M2: LSTM with LayerNorm regularization. (c) M3: Conv1D feature extraction followed by LSTM temporal modeling. (d) M4: TCN family with four dilated TCN blocks. (e) M5: Transformer encoder with Dense projection and two encoder blocks. (f) M6: residual LSTM with Add and LayerNorm before the final LSTM and TimeDistributed Dense layers.}
    \label{fig:modelarch}
\end{figure}

The six family architectures M1 to M6 and the three conditioning mechanisms (concat, film, gated) that constitute the 18-model comparison are described in Section~\ref{sec:s2s}; Figure~\ref{fig:sequence_layers} therein provides a component-level schematic of the main building blocks. Here,  Figure~\ref{fig:modelarch} shows all eighteen variants arranged by family architecture and conditioning mechanism.

\section{Tabor-Conditioning Mechanisms}\label{app:tabor_conditioning}

This appendix defines the three mechanisms used to inject the static Tabor parameter into the sequence-to-sequence models. Let ${\mathbf{X}\in \mathbb{R}^{N\times d}}$ denote the FMS-resampled input sequence, with $N=120$ measurement steps and $d$ sequential channels. In the main comparison $d=4$, corresponding to time, indentation, causal velocity, and the edge indicator. The physical Tabor parameter $\mu$ is first transformed and standardized using the same preprocessing pipeline as the other model inputs. We denote the resulting scalar network input by $z_{\mu}\in\mathbb{R}$. The surrogate therefore learns
\begin{equation}
f_{\theta}:\left(\mathbf{X},z_{\mu}\right)\mapsto \widehat{\mathbf{y}},
\qquad
\widehat{\mathbf{y}}\in\mathbb{R}^{N\times 1}.
\end{equation}
The three mechanisms differ only in where and how $z_{\mu}$ enters the neural architecture.

\begin{figure}[htbp]
    \centering
    \includegraphics[width=0.48\linewidth]{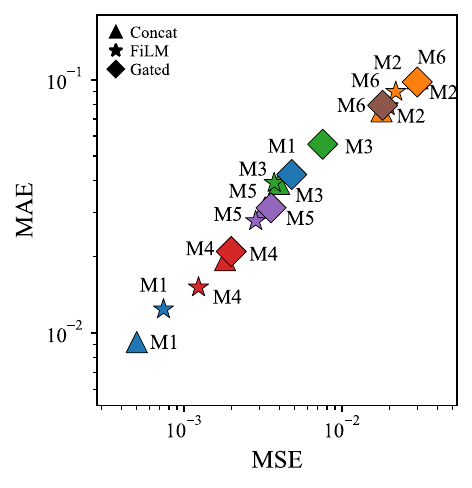}
    \caption{Held-out MSE versus MAE for the eighteen combinations of model family and Tabor-conditioning mechanism. Marker shapes identify concatenation, FiLM, and gated conditioning, while colors identify model families M1--M6.}
    \label{fig:tabor_conditioning_mse_mae}
\end{figure}

The comparison does not indicate a universal superiority of one conditioning. Based on the held-out MSE, concatenation performs best within M1 and M2, FiLM within M3--M5, and gated conditioning within M6. Thus, M1-concat is the best individual configuration, while the relative effectiveness of each conditioning mechanism depends on the temporal family and injection point.

\subsection{Concatenation conditioning}

Concatenation conditioning supplies the static scalar as an additional sequence channel. The scalar $z_{\mu}$ is repeated over the temporal dimension,
\begin{equation}
\mathbf{z}_{\mu}
=\left[
z_{\mu}, ...,
z_{\mu}\right]^\top
\in \mathbb{R}^{N\times 1},
\end{equation}
and appended to the sequence representation:
\begin{equation}
\widetilde{\mathbf{X}}
=
\left[\mathbf{X},\mathbf{z}_{\mu}\right]
\in \mathbb{R}^{N\times(d+1)}.
\end{equation}
The first temporal layer then receives $\widetilde{\mathbf{X}}$ rather than $\mathbf{X}$. This is an early-conditioning strategy: the recurrent, convolutional, or attention layer has access to the adhesion-regime parameter from the beginning of its sequence processing. This use of auxiliary variables through concatenated conditioning is part of the broader family of feature-wise conditioning operations discussed in \cite{dumoulin2018featurewise}.

\subsection{Feature-wise linear modulation conditioning}

Feature-wise Linear Modulation (FiLM) applies an affine transformation to hidden feature channels using coefficients generated from the conditioning variable \cite{perez2018film,dumoulin2018featurewise}. Let
\begin{equation}
\mathbf{H}=
\left[\mathbf{h}_1, ...,
\mathbf{h}_N\right]
\in \mathbb{R}^{N\times h}
\end{equation}
be the hidden sequence to be modulated. In the implementation, the conditioning path first maps $z_{\mu}$ through a dense layer with ReLU activation,
\begin{equation}
\mathbf{c}_{\mu}
=
\mathrm{ReLU}\!\left(\mathbf{W}_c z_{\mu}+\mathbf{b}_c\right)
\in \mathbb{R}^{64},
\end{equation}
and then produces a feature-wise scale vector and shift vector,
\begin{equation}
\boldsymbol{\gamma}_{\mu}
=
\mathbf{W}_{\gamma}\mathbf{c}_{\mu}+\mathbf{b}_{\gamma},
\qquad
\boldsymbol{\beta}_{\mu}
=
\mathbf{W}_{\beta}\mathbf{c}_{\mu}+\mathbf{b}_{\beta},
\qquad
\boldsymbol{\gamma}_{\mu},\boldsymbol{\beta}_{\mu}\in\mathbb{R}^{h}.
\end{equation}
The vectors are reshaped to $1\times h$ and broadcast over all measurement steps:
\begin{equation}
\widetilde{\mathbf{h}}_j
=
\boldsymbol{\gamma}_{\mu}\odot\mathbf{h}_j
+
\boldsymbol{\beta}_{\mu},
\qquad j=1,\ldots,N,
\end{equation}
or equivalently $\widetilde{\mathbf{H}}=\boldsymbol{\gamma}_{\mu}\odot\mathbf{H}+\boldsymbol{\beta}_{\mu}$ with temporal broadcasting. Because $z_{\mu}$ is constant for the whole trajectory, the same modulation is applied at every measurement step.

\subsection{Gated conditioning}

The gated variant uses the Tabor input to form a two-branch mixture. Let
\begin{equation}
\mathbf{A},\mathbf{B}\in\mathbb{R}^{N\times h}
\end{equation}
be two compatible temporal representations produced by parallel branches or by a representation and its normalized counterpart, depending on the family. The conditioning path maps $z_{\mu}$ to a hidden vector and then to two softmax weights:
\begin{equation}
\mathbf{q}_{\mu}
=
\mathrm{ReLU}\!\left(\mathbf{W}_g z_{\mu}+\mathbf{b}_g\right),
\qquad
\boldsymbol{\omega}_{\mu}
=
\mathrm{softmax}\!\left(\mathbf{W}_{\omega}\mathbf{q}_{\mu}+\mathbf{b}_{\omega}\right),
\end{equation}
with $\omega_{\mu,1}+\omega_{\mu,2}=1$. The fused hidden sequence is
\begin{equation}
\widetilde{\mathbf{H}}
=
\omega_{\mu,1}\mathbf{A}
+
\omega_{\mu,2}\mathbf{B}.
\end{equation}
The two scalar weights are broadcast over the temporal and feature dimensions. This mechanism is related to gated fusion models, where learned gates control the relative contribution of different representations \cite{arevalo2017gated}. In the present architecture, the gate is conditioned explicitly on the static Tabor input, so the mixture weights depend on the adhesion regime encoded by $\mu$.

\section{\textcolor{black}{Learning details and comparisons}}\label{sec:appendix_lc}
\begin{figure}
    \centering
    \includegraphics[width=0.49\linewidth]{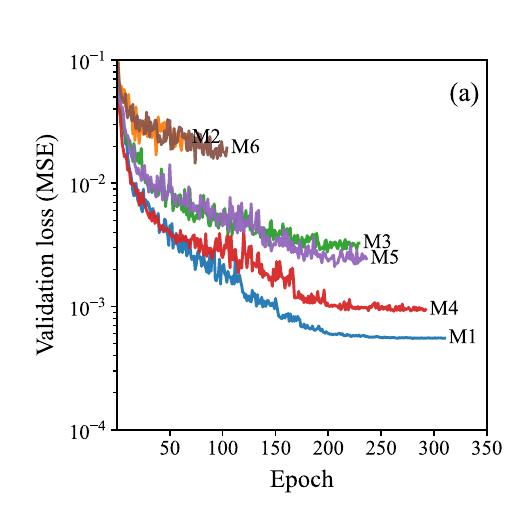}
    \hfill
    \includegraphics[width=0.49\linewidth]{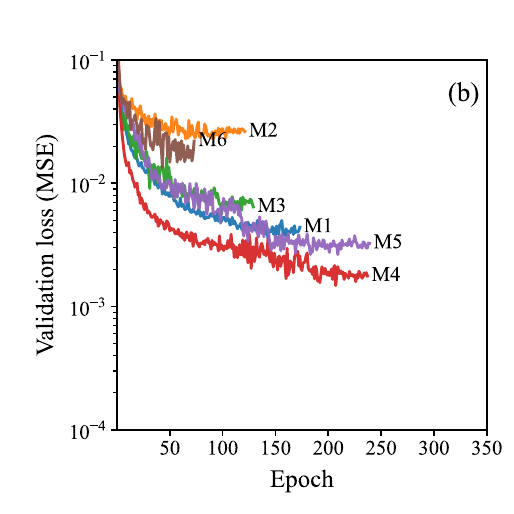}
    \caption{Validation-loss learning curves for the remaining conditioning mechanisms after moving the concatenation panel to Figure~\ref{fig:tier1d_pareto}a: \textbf{(a)} FiLM and \textbf{(b)} gated conditioning. Both panels use a shared epoch axis from 0 to 350 with matching tick locations, so convergence rates can be compared directly across conditioning mechanisms. None of the models shown required more than 350 epochs before early stopping.}
    \label{fig:tier1d_learning_curves_appendix}
\end{figure}

\textcolor{black}{\subsection{FiLM and gated conditioning learning curves}}
\textcolor{black}{This subsection} reports the validation-loss histories for the FiLM and gated conditioning mechanisms. Together with the concatenation curves in Figure~\ref{fig:tier1d_pareto}a, Figure~\ref{fig:tier1d_learning_curves_appendix} complements the summary metrics in the main text by showing the relative convergence rate, stability, and early-stopping behavior of each model during training.

\begingroup\color{black}
\subsection{Training-seed sensitivity}


To supplement the model comparison, we have chosen M1-concat and M4-concat. They were each trained with seeds 1337, 1339, 1341, and 1343 in Keras, keeping the data split, preprocessing, and optimization settings fixed. The seeds control initialization and stochastic training, including shuffling and dropout. Table~\ref{tab:r1_seed_variation} reports the mean and sample standard deviation of the four seed-level test means, using validation-selected checkpoints after early stopping. M1-concat has the smaller mean error for all four Table~\ref{tab:quartile_metrics} metrics, with an average pull-off-force error of $4.10\%$, compared with $7.91\%$ for M4-concat.
\endgroup

\begin{table}[htbp]
\centering\color{black}\small
\caption{\textcolor{black}{Training-seed sensitivity at $N=120$ with validation-based early stopping. Entries are the mean $\pm$ sample standard deviation across four seed-level test means.}}
\label{tab:r1_seed_variation}
\begin{tabular}{lcc}\toprule
Metric & M1-concat & M4-concat \\ \midrule
Force MSE ($\times10^{-3}$) & $0.556\pm0.051$ & $1.290\pm0.230$ \\
Pull-off force error (\%) & $4.10\pm0.35$ & $7.91\pm0.72$ \\
Hysteresis error (\%) & $2.54\pm0.10$ & $3.90\pm0.37$ \\
Pull-off time error (\%) & $0.59\pm0.06$ & $1.52\pm0.24$ \\
\bottomrule\end{tabular}
\end{table}

\begingroup\color{black}
\subsection{Non-sequential baseline}
To test whether force histories can be represented by a non-sequential regression model, XGBoost receives the complete flattened four-channel FMS input and the Tabor scalar, giving features at $N=120$, and fits one scalar regressor per force output. All regressors use the same frozen training/validation/test split and training-only normalization as the supplementary neural models. A 16-configuration grid varies the number of trees $\{100,300\}$, maximum depth $\{3,6\}$, learning rate $\{0.05,0.3\}$, and minimum child weight $\{1,10\}$. The regressors use seed 1337 and no early stopping. Configuration 9, with 300 trees, maximum depth 6, learning rate 0.3, and minimum child weight 10, is selected by the smallest validation FMS signed-log MSE and then evaluated once on the held-out test set. The mean $\pm$ standard deviation in Table~\ref{tab:r1_xgboost} are computed over the 1,245 held-out test trajectories from this selected configuration. Hysteresis uses the 935 complete observed loops with nonzero reference area. The matched seed-1337 M1-concat full-input model has a lower mean force MSE and smaller trajectory-to-trajectory standard deviation, $(0.540\pm1.022)\times10^{-3}$, than XGBoost, $(4.676\pm13.918)\times10^{-3}$.
\endgroup

\begin{table}[htbp]
\centering\color{black}\small
\caption{\textcolor{black}{Selected XGBoost non-sequential baseline. Entries are mean $\pm$ population standard deviation over the 1,245 held-out test trajectories for selected configuration 9.}}
\label{tab:r1_xgboost}
\setlength{\tabcolsep}{3pt}
\begin{tabular}{lcccc}\toprule
Model & \shortstack{Force MSE\\($\times10^{-3}$)} & \shortstack{Pull-off\\error (\%)} & \shortstack{Hysteresis\\error (\%)} & \shortstack{Pull-off time\\error (\%)} \\ \midrule
XGBoost & $4.676\pm13.918$ & $13.96\pm22.09$ & $3.13\pm7.85$ & $2.84\pm4.30$ \\
\bottomrule\end{tabular}
\end{table}

\section{BEM Discretisation Details}\label{app:BEM_disc}

\subsection{Kernel function}

The elastic Kernel function $G(r,s)$ appearing in Eq.~(\ref{eq:uz}) is:
\begin{equation}
G(r,s) = \begin{cases}
\dfrac{4}{\pi r}\,\overline{\mathrm{K}}\!\left(\dfrac{s}{r}\right), & s < r,\\[6pt]
\dfrac{4}{\pi s}\,\overline{\mathrm{K}}\!\left(\dfrac{r}{s}\right), & s > r,
\end{cases}
\label{eq:kernel}
\end{equation}
where $\overline{\mathrm{K}}(\cdot)$ is the complete elliptic integral of the first kind. Once the spatial grid is fixed, the influence matrix is assembled as $G_{ij} = G(r_i, r_j)\,r_j$, so that the elastic half-space deflection at node $i$ due to a traction field $\{\sigma_j\}$ is:
\begin{equation}
u_z[i] = \frac{1}{E_0^*}\sum_{j=1}^{N_s} G_{ij}\,\sigma_j.
\label{eq:uzdisc}
\end{equation}
The spatial distribution of traction on each element is approximated as piecewise linear using the \emph{method of overlapping triangles} \cite{johnson1987contact, papangelo2020numerical, papangelo2023detachment}.

\subsection{Dimensionless governing equations in discrete form}

The following dimensionless groups are introduced:
\begin{equation}
\widehat{h} = \frac{h - h_0}{h_0},\quad
\widehat{\delta} = \frac{\delta}{h_0},\quad
\widehat{r} = \frac{r}{\beta},\quad
\widehat{\sigma} = \frac{\sigma\,\mu\,h_0}{\Delta\gamma_0},\quad
\widehat{t} = \frac{t}{\tau_r},
\label{eq:nondim}
\end{equation}
where $\beta^3 = R^2\Delta\gamma_0/E_0^*$ is the characteristic contact half-width. With spatial index $i \in \{1,\dots,N_s\}$ and temporal index $q \in \{0,1,\dots\}$, the discretised Lennard-Jones law, gap equation, and displacement equation are:
\begin{equation}
\widehat{\sigma}[i,q] = -\frac{8}{3}\mu \left[\frac{1}{\bigl(\widehat{h}[i,q]+1\bigr)^{3}} - \frac{1}{\bigl(\widehat{h}[i,q]+1\bigr)^{9}}\right],
\label{eq:LJdisc}
\end{equation}
\begin{equation}
\widehat{h}[i,q] = -\widehat{\delta}[q] + \tfrac{1}{2}\mu\,\widehat{r}[i]^{2} + \widehat{u}_z[i,q],
\label{eq:hdisc}
\end{equation}
\begin{equation}
\widehat{u}_z[i,q] \approx \mu\sum_{j=1}^{N_s} \widehat{G}_{ij}\sum_{m=0}^{q} \widehat{C}[q-m]\,\bigl(\widehat{\sigma}[j,m+1] - \widehat{\sigma}[j,m]\bigr),
\label{eq:uzdisc2}
\end{equation}
where $\widehat{G}_{ij} = G_{ij}/(E_0^*\,\mu)$ is the dimensionless influence matrix, and the dimensionless SLS creep compliance is:
\begin{equation}
\widehat{C}(\widehat{t}) = 1 + (k-1)\exp(-\widehat{t}).
\label{eq:Cdisc}
\end{equation}
Equations~(\ref{eq:LJdisc})--(\ref{eq:Cdisc}) constitute a closed nonlinear system for $\widehat{h}[i,q]$ at each time step, solved by Newton-Raphson iteration at fixed $\widehat{\delta}[q]$.

\section{Neural Network Layer Definitions}\label{app:NN_components}

This appendix provides the complete mathematical definitions of the three building blocks used in the sequence-to-sequence architectures of Section~\ref{sec:s2s}.

\subsection{Long Short-Term Memory (LSTM) cell}

Let $\mathbf{x}_j \in \mathbb{R}^{d}$ be the input at step $j$, $\mathbf{h}_{j-1} \in \mathbb{R}^{H}$ the previous hidden state, and $\mathbf{c}_{j-1} \in \mathbb{R}^{H}$ the previous cell state. The LSTM cell update is \cite{hochreiter1997long}:
\begin{align}
\mathbf{f}_j &= \sigma_g\!\bigl(\mathbf{W}_f[\mathbf{h}_{j-1};\mathbf{x}_j]+\mathbf{b}_f\bigr), &
\mathbf{i}_j &= \sigma_g\!\bigl(\mathbf{W}_i[\mathbf{h}_{j-1};\mathbf{x}_j]+\mathbf{b}_i\bigr), \notag\\
\mathbf{o}_j &= \sigma_g\!\bigl(\mathbf{W}_o[\mathbf{h}_{j-1};\mathbf{x}_j]+\mathbf{b}_o\bigr), &
\tilde{\mathbf{c}}_j &= \tanh\!\bigl(\mathbf{W}_c[\mathbf{h}_{j-1};\mathbf{x}_j]+\mathbf{b}_c\bigr), \label{eq:lstm_app}\\
\mathbf{c}_j &= \mathbf{f}_j\odot\mathbf{c}_{j-1}+\mathbf{i}_j\odot\tilde{\mathbf{c}}_j, &
\mathbf{h}_j &= \mathbf{o}_j\odot\tanh(\mathbf{c}_j), \notag
\end{align}
where $\sigma_g$ is the sigmoid activation, $\odot$ is the Hadamard product, and $[\cdot;\cdot]$ denotes concatenation. Matrices $\mathbf{W}_{\cdot}\in\mathbb{R}^{H\times(H+d)}$ and biases $\mathbf{b}_{\cdot}\in\mathbb{R}^{H}$ are learnable. The forget gate $\mathbf{f}_j$ regulates how much prior cell information is retained, while the full recurrent update enables the LSTM to represent the fading-memory behavior associated with the Boltzmann convolution of Eq.~(\ref{eq:uz}). In the M1 concatenation and FiLM variants, two LSTM layers are chained: $H_1 = 256$ hidden units in the first layer and $H_2 = 128$ in the second, so that the recurrent family has $4H_1(H_1+d_{\mathrm{in}}+1) + 4H_2(H_2+H_1+1)$ parameters, where $d_{\mathrm{in}}$ is the feature dimension seen by the first recurrent layer. For the physics-guided representation, $d_{\mathrm{in}}=4$ for the FiLM variant and $d_{\mathrm{in}}=5$ for the concatenation variant, yielding 464{,}384 and 465{,}408 recurrent parameters, respectively, before the readout. These counts do not apply to M1-gated, because that implementation uses two parallel LSTM(128) expert branches and a softmax fusion gate rather than the same 256$\to$128 recurrent family. With the shared readout of Eq.~(\ref{eq:tdd_app}), the M1-concat total is therefore $465{,}408+8{,}321=473{,}729$ trainable parameters. The M1-FiLM total is $464{,}384+16{,}768+8{,}321=489{,}473$, where 16{,}768 parameters come from the FiLM conditioning network, and the two-expert M1-gated implementation contains 144{,}899 trainable parameters. These are the corresponding M1 values used in Figure~\ref{fig:tier1d_pareto}(b).

\subsection{Temporal Convolutional Network (TCN) block}

A residual TCN block at layer $l$ with dilation $d_l = 2^{l-1}$, kernel size $K$, and $C_l$ output channels applies two causal convolutions before adding the skip path \cite{bai2018empirical}:
\begin{align}
\tilde{\mathbf{z}}_j^{(l,1)} &= \mathrm{ReLU}\!\left(\mathrm{LayerNorm}\!\left(\sum_{s=0}^{K-1} \mathbf{W}_{1,s}^{(l)}\,\mathbf{z}_{j-d_l s}^{(l-1)} + \mathbf{b}_1^{(l)}\right)\right), \notag\\
\tilde{\mathbf{z}}_j^{(l,2)} &= \mathrm{ReLU}\!\left(\mathrm{LayerNorm}\!\left(\sum_{s=0}^{K-1} \mathbf{W}_{2,s}^{(l)}\,\tilde{\mathbf{z}}_{j-d_l s}^{(l,1)} + \mathbf{b}_2^{(l)}\right)\right), \notag\\
\mathbf{z}_j^{(l)} &= \tilde{\mathbf{z}}_j^{(l,2)} + \mathbf{P}^{(l)}\mathbf{z}_j^{(l-1)},
\label{eq:tcn_app}
\end{align}
where left padding enforces causality by treating indices $q<1$ as zero, dropout (rate 0.1, omitted from the notation) is applied after each activation in the implemented block, and $\mathbf{P}^{(l)}$ is the identity when the channel width is unchanged and a learned pointwise projection otherwise. With $L = 4$ blocks, $K = 3$, and dilations $d_l = 1,2,4,8$, the receptive field of the M4 architecture is:
\begin{equation}
\mathcal{R} = 1 + 2(K-1)\sum_{l=1}^{L} d_l = 1 + 2(K-1)(2^L - 1) = 1 + 4\cdot15 = 61 \text{ steps},
\end{equation}
covering approximately one-half of the 120-step sequence without any recurrent state. For a constant-width TCN with $C$ channels, the family scales as $\mathcal{O}(2L\cdot K\cdot C^2)$ because each residual block contains two causal convolutions. For the implemented M4 widths $4\to64\to64\to128\to128$, the four TCN blocks contain $13{,}760+24{,}960+82{,}816+99{,}072=220{,}608$ trainable parameters, including convolution biases, LayerNorm scale and shift parameters, and the learned pointwise residual projections when the channel width changes. After Tabor conditioning and the TimeDistributed readout, the trainable-parameter totals used in Figure~\ref{fig:tier1d_pareto}(b) are 228{,}993 (M4-concat), 245{,}697 (M4-FiLM), and 229{,}315 (M4-gated).

\subsection{TimeDistributed Dense layer}
{The implemented readout maps the stepwise latent representation $\mathbf{z}_j \in \mathbb{R}^{H}$ to a scalar force prediction through two shared time-distributed dense transformations. First, a 64-unit ReLU layer is applied independently at each measurement step, followed by a shared scalar affine projection \cite{bai2018empirical}:}
\begin{equation}
\mathbf{r}_j = \mathrm{ReLU}\!\left(\mathbf{W}_r\mathbf{z}_j+\mathbf{b}_r\right),
\qquad
\widehat{P}_j = \mathbf{w}_o^\top\mathbf{r}_j + b_o,
\quad j = 1,\dots,N,
\label{eq:tdd_app}
\end{equation}
where $\mathbf{W}_r\in\mathbb{R}^{64\times H}$, $\mathbf{b}_r\in\mathbb{R}^{64}$, $\mathbf{w}_o\in\mathbb{R}^{64}$, and $b_o\in\mathbb{R}$ are shared across all time steps. The readout therefore adds $64(H+1)+65$ trainable parameters independent of the sequence length $N$. For M1 with $H_2 = 128$, this corresponds to 8{,}321 readout parameters.

\section{\textcolor{black}{Effects of Training-Set and Physics-Guided Input Representation}}\label{app:feature_comparison}

\subsection{Physics-Guided Input representation effect}

\begin{figure}[htbp]
    \centering
    \includegraphics[width=0.49\linewidth]{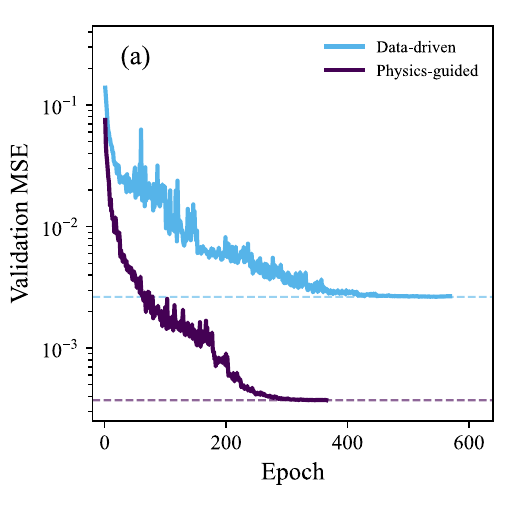}
    \hfill
    \includegraphics[width=0.49\linewidth]{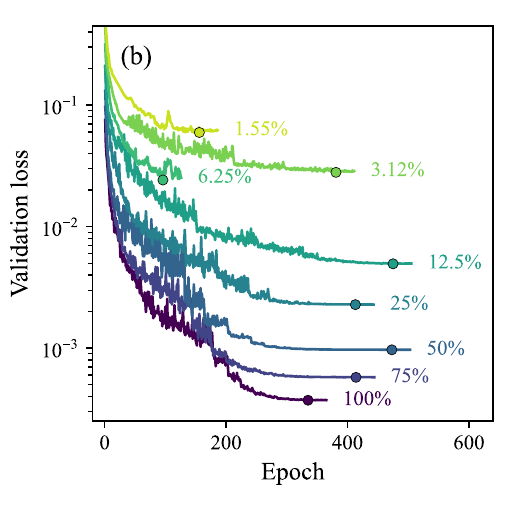}
    \caption{Effects of input representation and training-set size on M1-concat optimization. \textbf{(a)} Validation-MSE learning curves for the data-driven representation (time and indentation only) and the physics-guided representation (augmented with causal velocity and edge detection), trained with identical hyperparameters and the same 80/10/10 split. \textbf{(b)} Validation-loss histories for the paper M1-concat model trained on the full partition and for models trained on nested fractions of that partition; circles mark the minimum validation loss of each run. Both panels use logarithmic loss axes.}
    \label{fig:physics_informed_vs_data_driven_AD}
\end{figure}

Figure~\ref{fig:physics_informed_vs_data_driven_AD}(a) isolates the role of feature design by comparing two versions of the same M1-concat model. The data-driven representation receives only time and indentation, whereas the physics-guided representation augments those inputs with causal velocity and an edge-detection feature that highlights loading-history transitions. Under identical training, validation, and test splits, the physics-guided representation converges more quickly and reaches a best validation MSE approximately seven times lower than the data-driven alternative. \RevisionBlue{This improvement is best interpreted primarily as a feature-engineering benefit. Causal velocity makes the rate history explicit, while the edge channel flags abrupt velocity changes associated with the prescribed loading--dwell--unloading transitions. Both are deterministic transforms of the supplied time--indentation history, so they reorganize existing protocol information rather than add new measured physical information.}

\RevisionBlue{An LSTM can, in principle, infer loading and unloading phases and their rates from the ordered time--indentation sequence, since these are encoded in its temporal increments. Figure~\ref{fig:physics_informed_vs_data_driven_AD}(a) compares only the two-channel input with the four-channel input, adding velocity and edge together.}

\subsection{\textcolor{black}{Data availability and computational costs}}\label{app:data_costs}

To separately quantify the effect of data availability, the manuscript M1-concat architecture was trained on nested subsets containing 75\%, 50\%, 25\%, 12.5\%, 6.25\%, 3.125\%, and 1.55125\% of the original training partition. For each fraction, preprocessing statistics were fitted using only the available training subset, whereas the validation and test indices were held fixed. This construction prevents information leakage and ensures that the comparison probes reduced training coverage rather than a changing evaluation set. Figure~\ref{fig:physics_informed_vs_data_driven_AD}(b) shows that reducing the number of trajectories raises the validation-loss floor and generally increases optimization variability. Moderate reductions retain the overall convergence pattern of the full-data model, whereas the smallest subsets plateau at substantially larger losses.

\textcolor{black}{For $n_q$ subsequent trajectory queries, the amortized costs are $C_{\mathrm{BEM}}(n_q)=n_q\bar t_{\mathrm{BEM}}$ for direct BEM and $C_{\mathrm{DL}}(n_q)=C_{\mathrm{offline}}+n_q\bar t_{\mathrm{DL}}$ for the surrogate. Here, $\bar t_{\mathrm{BEM}}$ is the mean BEM runtime per trajectory, while $\bar t_{\mathrm{DL}}$ is the mean batch-one inference time to predict one complete force trajectory, averaged over repeated timings of the evaluated protocols. The offline cost is $C_{\mathrm{offline}}=C_{\mathrm{data}}+C_{\mathrm{fit}}+C_{\mathrm{other}}$. Where $C_{\mathrm{data}}$ is the cost of generating the trajectories used for training (and the fixed validation and test sets); $C_{\mathrm{fit}}$ is the cost of fitting the final surrogate once; and $C_{\mathrm{other}}$ covers any additional model-selection and development costs not already included in those terms. In the plotted illustrative scenario, $C_{\mathrm{other}}$ is set to zero. The BEM database required approximately 478 single-core CPU-hours for \AM{12,446} trajectories (about $138\,\mathrm{s}$ per trajectory), and the full-data M1-concat fit required 105,005 s ($29.2\,\mathrm{h}$). These are approximate timings on different bases. This means BEM CPU-hours are treated as serial elapsed time, while the fit time is the recorded workstation wall time.}

\begingroup\color{black}
For Figure~\ref{fig:accuracy_amortization}, fit times for reduced training fractions were estimated by scaling the measured reference-fit time according to the recorded stopping epochs and training and validation batch counts. The corresponding BEM database-generation cost was added, retaining the fixed validation and test sets. The full-data setting was charged for all \AM{12,446} trajectories. The figure relates these estimated costs to the observed mean test MSE for each training fraction.

In panel~(a) of Figure~\ref{fig:accuracy_amortization}, the black line shows estimated cumulative BEM cost against subsequent queries, while each colored surrogate-cost line begins at its break-even point. The legend gives mean test MSE and training fraction. Panel~(b) shows the estimated break-even query count on the horizontal axis and the observed mean test MSE, treated as an acceptable-error threshold, on the logarithmic vertical axis for the eight training fractions. The curve serves as a visual guide.

Under the stated cost assumptions, the rose region identifies combinations of query volume and acceptable mean test error for which direct BEM has the lower estimated total cost. In the blue region, the upfront simulation and training costs have been amortized, making our DL+BEM framework more cost-effective. Direct numerical simulation is therefore preferable on computational-cost grounds when only a small number of trajectories is required, especially when a strict accuracy target would be required for every single simulation. For applications requiring many repeated predictions over extended operating periods, DL+BEM can recover its initial investment and provide faster responses, provided the operating conditions remain within the validated domain and its prediction errors meet the application requirements. Among the tested fractions, tighter mean-error targets are associated with larger offline costs and higher break-even query counts.

\begin{figure}[H]
    \centering
    \includegraphics[width=0.49\linewidth]{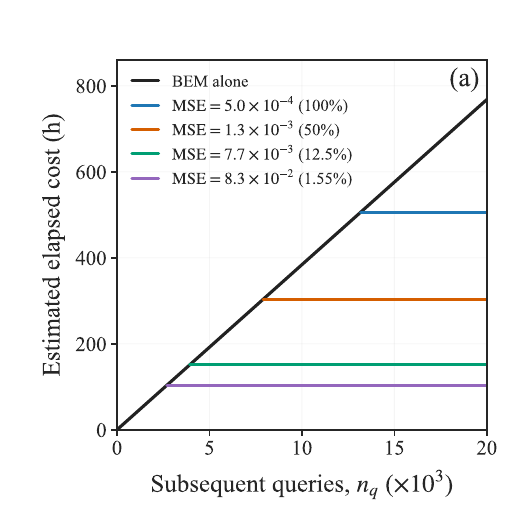}
    \hfill
    \includegraphics[width=0.49\linewidth]{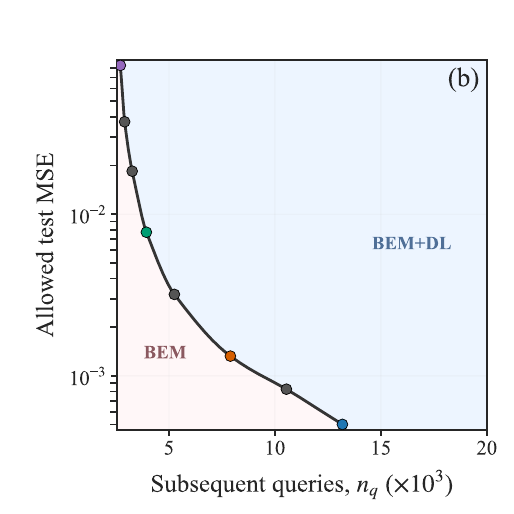}
    \caption[\textcolor{black}{Accuracy and query volume in the amortized-cost comparison}]{\RevisionBlue{\input{figures/generated/figure_G2_amortization.caption.tex}}}
    \label{fig:accuracy_amortization}
\end{figure}
\endgroup

\begingroup\color{black}
\subsection{Interpolation and extrapolation across withheld Tabor bands}
To study the effect of unseen regions of exploration space. The eligible native dataset is divided into four contiguous Tabor bands with approximately equal populations, keeping identical $\mu$ values together. We consider two holdout tests using the interior band B2 and the lowest band B1. The allocation of the data in these two tests is summarized in Table~\ref{tab:r1_band_partitions}. In the B1 holdout, the test set is the complete lowest band, B1, while training and validation use trajectories from B2, B3, and B4. Thus, the smallest Tabor parameter seen during training or validation is $\mu=0.4000$, whereas the test range extends down to $\mu=0.2000$. In the B2 holdout, the complete interior band B2 is reserved for testing. Training and validation contain B1 below the gap and B3--B4 above it, so the test range is bounded by Tabor values represented on both sides of the retained data.

\begin{table}[htbp]
\centering\color{black}\small
\caption{\textcolor{black}{Tabor-band composition of the interpolation and extrapolation holdout tests.}}
\label{tab:r1_band_partitions}
\setlength{\tabcolsep}{4pt}
\begin{tabular}{llccc}
\toprule
Experiment & Partition & Bands used & Cases & $\mu$ range \\
\midrule
B1 holdout & Training & B2+B3+B4 & 8,160 & $0.4000$--$3.2000$ \\
           & Validation & B2+B3+B4 & 1,021 & $0.4000$--$3.2000$ \\
           & Test & B1 & 3,265 & $0.2000$--$0.3311$ \\
\addlinespace
B2 holdout & Training & B1+B3+B4 & 8,172 & $0.2000$--$3.2000$ \\
           & Validation & B1+B3+B4 & 1,022 & $0.2000$--$3.2000$ \\
           & Test & B2 & 3,252 & $0.4000$--$0.7053$ \\
\bottomrule
\end{tabular}
\end{table}

Table~\ref{tab:r1_holdouts} shows the results of the B1 extrapolation and B2 interpolation tests. The results show larger errors for the B1 extrapolation test than for the B2 interpolation test. The mean force MSE is $0.612\times10^{-3}$ for B2 and $5.832\times10^{-3}$ for B1. Thus, interpolation across the interior gap retains more accuracy than extrapolation below the retained training range in these tests. 
\endgroup

\begin{table}[H]
\centering\color{black}\small
\caption{\textcolor{black}{M1-concat generalization under random and Tabor-band holdouts using the Table~\ref{tab:quartile_metrics} FMS metric definitions. Entries are mean $\pm$ standard deviation across the indicated test trajectories.}}
\label{tab:r1_holdouts}
\setlength{\tabcolsep}{3pt}
\begin{tabular}{lrcccc}\toprule
Test & Cases & \shortstack{Force MSE\\($\times10^{-3}$)} & \shortstack{Pull-off\\error (\%)} & \shortstack{Hysteresis\\error (\%)} & \shortstack{Pull-off time\\error (\%)} \\ \midrule
Random & 1,245 & $0.540\pm1.022$ & $4.28\pm6.63$ & $2.59\pm6.38$ & $0.59\pm1.29$ \\
B2: interpolation & 3,252 & $0.612\pm0.874$ & $4.15\pm4.48$ & $2.88\pm3.74$ & $0.32\pm0.82$ \\
B1: extrapolation & 3,265 & $5.832\pm12.849$ & $9.36\pm37.47$ & $8.52\pm8.99$ & $0.59\pm1.36$ \\
\bottomrule\end{tabular}
\end{table}

\section{\textcolor{black}{FMS-resolution and prediction errors}}\label{app:fms_resolution_transfer}

Recurrent neural-network layers such as LSTMs are not tied, at the level of their recurrent weights, to one fixed number of sequence steps. The same recurrent cell is applied successively along the sequence dimension, as schematically shown in Figure~\ref{fig:sequence_layers}(a). In the present surrogate, however, physical time is itself one of the input channels. Therefore, the model should not be interpreted as being agnostic to physical time. Rather, the relevant architectural property is flexibility with respect to the number of fixed-measurement-step (FMS) points used to represent a given loading history.

\begin{figure}[htbp]
    \centering
    \includegraphics[width=0.49\linewidth]{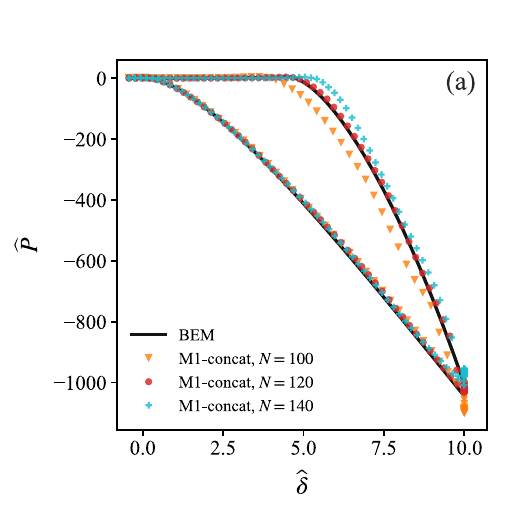}
    \hfill
    \includegraphics[width=0.49\linewidth]{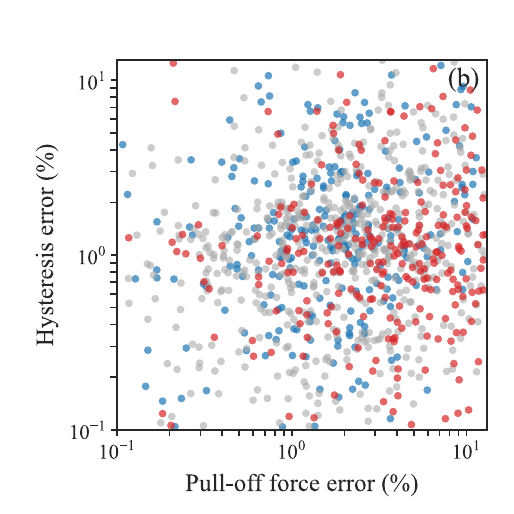}
    \caption[\textcolor{black}{FMS-resolution and prediction errors}]{\RevisionBlue{\input{figures/generated/figure_H12.caption.tex}}}
    \label{fig:fms_resolution_transfer}
\end{figure}

To examine this point, the trained M1-concat model, selected using the reference resolution $N=120$, was evaluated on the same representative BEM trajectory using nearby FMS resolutions $N=100$ and $N=140$, without retraining. Figure~\ref{fig:fms_resolution_transfer}(a) shows that the best agreement is obtained at the training resolution $N=120$, as expected. The predictions obtained with $N=100$ and $N=140$ still follow the overall force--indentation trend, but with visible deviations, indicating that modest changes in FMS resolution can be processed by the trained recurrent model while the accuracy remains resolution-dependent. This provides a practical advantage of the selected recurrent surrogate: it is not hard-restricted to the exact FMS grid used during training, so nearby resolutions can be exploited without changing the network architecture or retraining the weights.

\RevisionBlue{Figure~\ref{fig:fms_resolution_transfer}(b) examines prediction errors at the extremes of the unloading-rate distribution by recolouring the pull-off-force versus hysteresis-error comparison in Figure~\ref{fig:error_distribution}(c). The slowest and fastest unloading-rate quartiles are highlighted in blue and red, respectively, while the middle two quartiles are grey. This comparison distinguishes local detachment-force accuracy from the integrated hysteresis error; neither metric alone resolves the detailed force evolution during dwell relaxation or the rapid force change at snap-off.}

\begingroup\color{black}
To complement the single-trajectory transfer test, separate M1-concat models are trained at $N=90$, $120$, and $150$ using seed 1337, the same 9,956/1,245/1,245 protocol split, and validation-based early stopping. The loading/dwell/unloading allocations for this investigation are $(37,16,37)$, $(50,20,50)$, and $(62,26,62)$, respectively. Here, each model is trained and tested at its own resolution. All physical errors in Table~\ref{tab:r1_native_resolution} use the original MATLAB BEM force histories before FMS resampling as the reference. Predictions are mapped back to the observed native time grid, and their maximum tensile unloading force is compared with the native BEM maximum. Thus, this comparison includes the effect of FMS representation on the prediction error instead of measuring pull-off only relative to a potentially smoothed FMS target. Increasing $N$ from 90 to 150 reduces mean native-reference pull-off error from $6.34\%$ to $4.88\%$ and mean pull-off-time error from $1.39\%$ to $1.05\%$ of the prescribed unloading duration.
\endgroup

\begin{table}[htbp]
\centering\color{black}\small
\caption{\textcolor{black}{Native-reference physical-event errors after separate training at each FMS resolution. Entries are mean $\pm$ sample standard deviation across the common test population.}}
\label{tab:r1_native_resolution}
\setlength{\tabcolsep}{5pt}
\begin{threeparttable}
\begin{tabular}{rccc}\toprule
$N$ & Pull-off force w.r.t. native (\%) & Pull-off time w.r.t. native (\%) & Hysteresis w.r.t. native (\%) \\ \midrule
90 & $6.34\pm9.57$ & $1.39\pm1.68$ & $2.84\pm5.35$ \\
120 & $5.76\pm8.23$ & $1.21\pm1.44$ & $2.67\pm6.52$ \\
150 & $4.88\pm6.66$ & $1.05\pm1.32$ & $2.21\pm4.05$ \\
\bottomrule\end{tabular}
\begin{tablenotes}[flushleft]\footnotesize
\item \textit{$N$}: number of FMS points; a separate M1-concat model is trained and tested at each resolution. \textit{Pull-off force w.r.t. native}: relative error between the predicted and native-BEM maximum tensile forces during unloading.  \textit{Pull-off time w.r.t. native}: absolute event-time error normalized by the prescribed unloading duration. \textit{Hysteresis w.r.t. native}: relative error between absolute force--indentation loop areas, evaluated for complete native-BEM loops with nonzero reference area.
\end{tablenotes}
\end{threeparttable}
\end{table}

\end{document}